\documentclass[manuscript,screen]{acmart}
\usepackage[edges]{forest}
\usepackage{adjustbox}
\usepackage{graphicx}
\graphicspath{{images/}}
\usepackage{subfiles}
\usepackage{subcaption}
\usepackage{amsmath}
\usepackage[usestackEOL]{stackengine}

\AtBeginDocument{%
  }

\setcopyright{acmcopyright}
\copyrightyear{2024}
\acmJournal{CSUR}
\acmYear{2024}

\begin{document}

\title{A Survey of Deep Long-Tail Classification Advancements}

\author{Charika De Alvis}
\affiliation{%
  \institution{The University of Sydney}
  \city{Sydney}
  \country{Australia}}
\email{charika.weerasiriwardhane@sydney.edu.au}

\author{Suranga Seneviratne}
\affiliation{%
  \institution{The University of Sydney}
  \city{Sydney}
  \country{Australia}}
\email{suranga.seneviratne@sydney.edu.au}

\begin{abstract}
Many data distributions in the real world are hardly uniform. Instead, skewed and long-tailed distributions of various kinds
are commonly observed. This poses an interesting problem for machine learning, where most
algorithms assume or work well with uniformly distributed data. The problem is further exacerbated by current state-of-the-art deep
learning models requiring large volumes of training data. As such, learning from imbalanced data remains a
challenging research problem and a problem that must be solved as we move towards more real-world applications of deep learning. In the context of class imbalance, state-of-the-art (SOTA) accuracies on standard benchmark datasets for classification typically fall
less than 75\%, even for less challenging datasets such as CIFAR100. Nonetheless, there has been progress in this niche
area of deep learning. To this end, in this survey, we provide a taxonomy of various methods proposed for addressing the problem of
long-tail classification, focusing on works that happened in the last few years under a single mathematical framework.
We also discuss standard performance metrics, convergence studies, feature distribution and classifier analysis. We also provide a quantitative comparison of the performance of different SOTA methods and conclude the survey by discussing the remaining challenges and future research direction.
\end{abstract}

\ccsdesc[500]{Computing methodologies~Supervised learning by classification}

\keywords{supervised learning, deep learning, long-tailed classification, class imbalance}

\received{22 April 2007}

\maketitle

\section{Introduction}

\begin{figure}[h]
\begin{subfigure}{.45\textwidth}
\includegraphics[width=0.9\linewidth,height=0.7\linewidth]{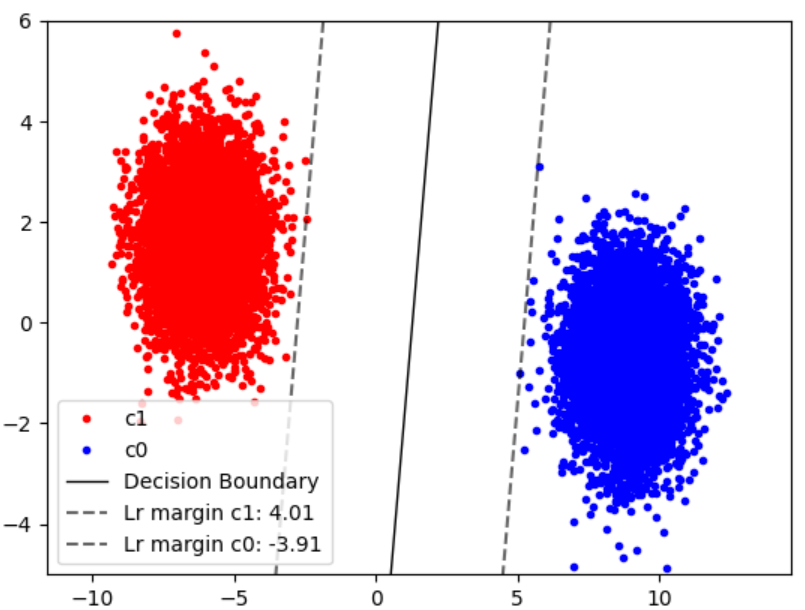}  
  \caption{Logistic Regression - Rlass imbalance ratio 1:1}
  \label{fig:1}
\end{subfigure}
\begin{subfigure}{.45\textwidth}
\includegraphics[width=0.9\linewidth,height=0.7\linewidth]{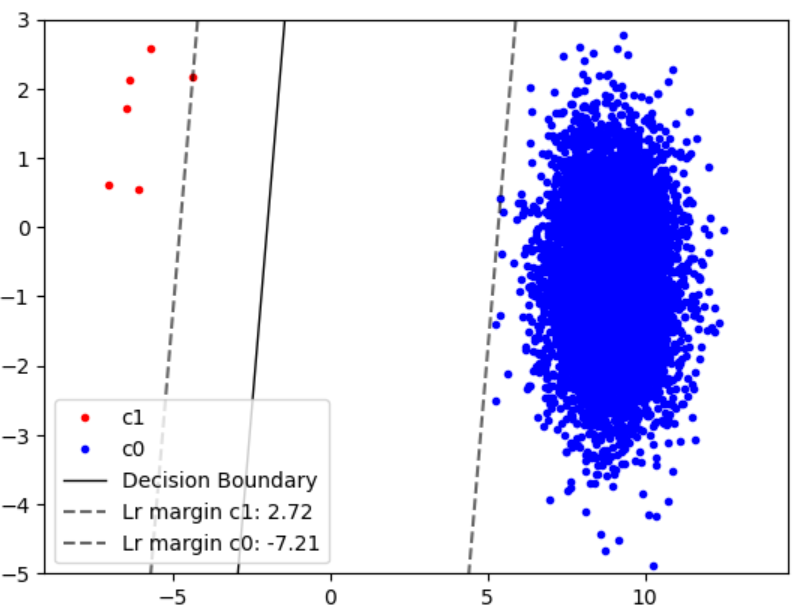}  
  \caption{Logistic Regression - Class imbalance ratio 1:1000}
  \label{fig:sub-first}
\end{subfigure}

\caption{Denotes the drift of the decision boundary towards the minority class boundary with the increasing class imbalance ratio.}
\label{fig:logistic}
\end{figure}

Over the last decade, deep learning has become the de-facto method of gaining value from big data in computer vision, natural language processing, speech recognition, and many other application domains. While deep learning methods have outperformed humans as well as traditional methods in
many learning tasks, they are not without weaknesses. One such weakness that
limits using deep learning methods in some of the critical real-world applications is their inability to handle highly imbalanced data. In training data, some classes tend to have a significantly larger number of samples compared to the other classes causing a long-tailed distribution. Examples of real-world applications with inherently imbalanced data distributions include fraud detection~\cite{fraud}, spam detection~\cite{spam}, pedestrian detection~\cite{pedes}, and network intrusion detection~\cite{intrusion}. Machine learning in such settings, traditional machine learning or deep learning, creates an inherent bias towards majority classes during training. For instance, in Figure~\ref{fig:logistic}, we show an example of binary classification using logistic regression on linearly separable data. As can be seen, when the class imbalance ratio increases, the class margin for the minority class grows thinner. In other words, the minority class tend to fit the data, leaving insufficient generalization. In related applications, this poses a challenge. For instance, in fraud detection, where fraudulent transactions are the minority, failing to identify a fraudulent transaction carries more severe repercussions than mistakenly flagging a benign transaction as fraudulent.

There has been extensive research on dealing with class imbalance using traditional machine learning methods.  One such approach is to address the bias towards the majority class by adjusting the training data to reduce the imbalance (i.e., data-level solutions)~\cite{126}, \cite{127}, \cite{128}, \cite{129} and ~\cite{root}. Another option is to modify the model's underlying learning or decision-making process to enhance its sensitivity towards the minority classes (., algorithmic level solutions) \cite{veropoulos1999controlling} 
\cite{iranmehr2019cost}
\cite{fan2017learning}. In deep learning models, the issue of class imbalance becomes more pronounced, as the majority class heavily influences the net gradient responsible for updating the model's weights~\cite{survey11}. Therefore, the error of the majority group decreases rapidly during the early iterations. However, this phenomenon also leads to an increased error for the minority class, causing the network to become trapped in a slow convergence mode.

There are surveys, such as~\cite{survey9}, \cite{survey8}, \cite{survey7}, and \cite{survey10}, that discuss practical issues associated with class imbalance in conventional machine learning and present existing solutions under three categories; data-level, algorithmic-level, and hybrid solutions. Under data-level solutions, the most discussed topics are sampling, bagging and boosting.  Typically under algorithmic level solutions, cost-sensitive decision trees, cost-sensitive SVMs and cost-sensitive neural networks were analysed. Under hybrid methods, means for efficient combination of the fundamental data-level and algorithmic-level solutions is discussed. Fundamental performance evaluation metrics for class imbalance settings such as ROC curves, and confusion metrics were also demonstrated with illustrations.  However, these surveys mostly discuss non-deep learning and fundamental deep-learning methods.

On the other hand, some of the previous surveys of deep long-tail (LT) classification~\cite{survey11}, \cite{survey12}, \cite{survey13}, \cite{survey14} do not cover recent advancements and novel model evaluation ideas. Nonetheless, there are several recent surveys conducted on long-tail classification. For example, in~\cite{s1} the authors discuss a chronological overview of the LT classification methods.  
Chaowei et al.~\cite{s2} evaluate and compare the SOTA LT methods based on a new metric that computes the accuracy of the classification, stability and bounds for dynamically evolving test data distributions. Another survey specific to visual recognition~\cite {s3} provides theoretical explanations for the different SOTA LT classification methods. 
Finally, Yifan et al.~\cite{s4} introduce a taxonomy for SOTA LT classification methods highlighting three main categories; class re-balancing, information augmentation, and module improvement. They use a new metric termed relative accuracy to obtain an empirical understanding of the extent to which different SOTA methods contribute to alleviating the imbalance of the training data. Nonetheless, most of these surveys lack an explanation of various deep LT techniques in a common notation so that the connection between different methods can be clearly established. In addition, the discussion of model performance techniques is confined to a narrow scope. Finally, they require a broad analysis of the research gaps, LTR trends and important future research directions.

To this end, in this survey, we provide a detailed review of the existing deep LT classification methods while highlighting the intuitions for the methods, and inter-dependencies of different techniques under a unified mathematical framework. In addition, we discuss the model performance analysis for LT tasks going beyond standard metrics. We also discuss long-tail ness in online settings and the benefits of zero-shot learning to LTR. More specifically, we make the following contributions. 
\subsection{Main Contributions of the Survey }
\begin{itemize}
    \item  We provide a taxonomy for SOTA algorithmic level solutions to the deep  classification problem consisting of four main branches: Loss Modification, Margin-based Loss reweighting, Representation Learning Improvements, and Balanced Classifier Learning.

\item We use a common mathematical notation to describe the methods included in the taxonomy so that the intuition behind the methods, interconnections, and dependencies can be interpreted clearly.
\item  We showcase efficient metrics and strategies for evaluating and comparing SOTA algorithms under four main aspects: standard metrics,  convergence studies, classifier analysis, and feature distribution analysis.

\item Finally, we discuss existing challenges, research gaps and possible future directions in deep long-tail classification in general and specifically in the areas of online learning and zero-shot learning

\end{itemize}

The rest of our survey is organised as follows. In Section 2, we present the background information, including the definition of the classification problem, commonly used performance metrics, and datasets. Section 3 introduces our overall taxonomy for classifying algorithm-level solutions for deep long-tail classification. Section 4 describes the Loss Reweighting methods while Section 5 describes Margin-based loss modification methods. In Section 6 we describe Optimized representation learning methods and in Section 7 we describe Balanced classifier learning methods. We discuss the existing challenges and future research directions in Section 8, and finally, Section 9 concludes the paper.

\section{Background}

We start by defining the long-tail classification problem and associated performance metrics. Afterwards, we describe the commonly used datasets and experiment settings using the works we cover in this survey.

\subsection{Problem Definition}
\label{problem}
   Multiclass classification problem can be defined by input instance domain $\mathbb{X}$, class label domain is given by $ Y=[L]=\{1,……,L\}$. For a set of training samples $ S = \{(x_i,y_i)\}_{i=1}^N \tilde P^N  $  where joint distribution $P$ characterized on $\mathbb{X} \times Y$. The multi-class learning task is to learn a scorer $f: \mathbb{X} \to R^L$ so that it minimizes the expected loss $l: Y \times R^L \to R_+$, which is equivalent to solving the following optimization problem. Here, $R_+$ denotes the set of positive real numbers.

\begin{equation}
\text{min}_{f \in F} \mathbb{R}(f) = E_{(x,y) \tilde P} [ l(y,f(x))]
\end{equation}

Where $F$ indicates a class of models for the scorer. Misclassification error is denoted by $\mathbb{R}(f)$.  $l(y,f(x)$ can be resembled by the zero-one loss but as it is not differentiable, a surrogate loss is used in model optimization. Common practice is to use the SoftMax cross entropy loss denoted in Equation~\ref{ce1}.

\begin{equation}
 l_{CE}(y,f(x)) = - \text{log} \frac{e^{f_y(x)}}{\sum_{{y’} \in Y} e^{f_{y’}(x)}} =\text{log} [1+ \sum_{y’ \neq y} e^{f_{y’}(x)-f_y(x)}]
 \label{ce1}
\end{equation}

\begin{figure}[h]

\hspace{-6mm}
\centering
\begin{subfigure}{.4\textwidth}
\includegraphics[width=0.7\linewidth]{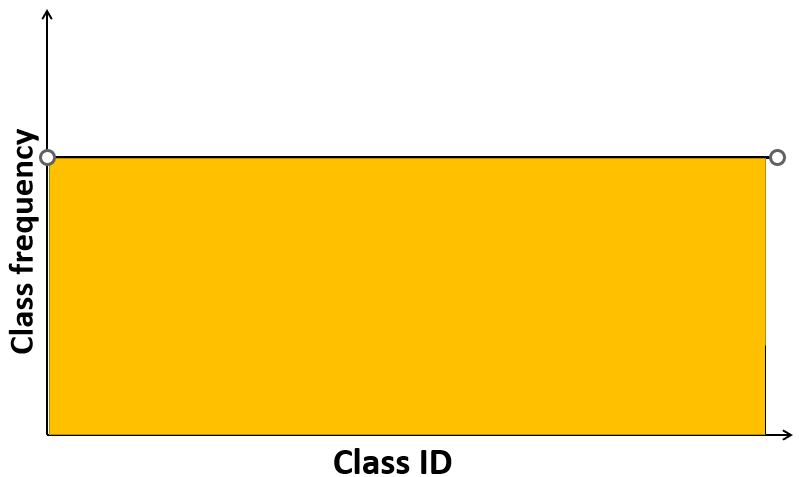}  
  \caption{Uniform Distribution}
  \label{fig:sub-first}
\end{subfigure}
\hspace{-18mm}
\begin{subfigure}{.4\textwidth}
\includegraphics[width=0.7\linewidth]{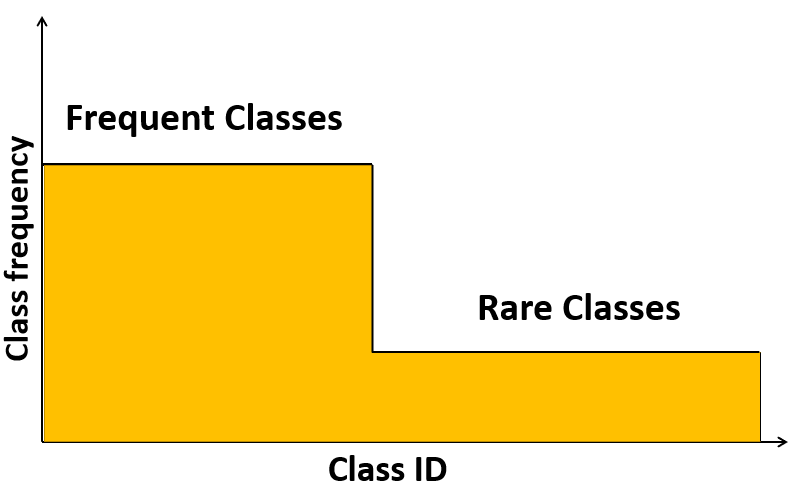}  
  \caption{Step Distribution}
  \label{fig:sub-first}
\end{subfigure}
\hspace{-18mm}
\begin{subfigure}{.4\textwidth}
\includegraphics[width=0.7\linewidth]{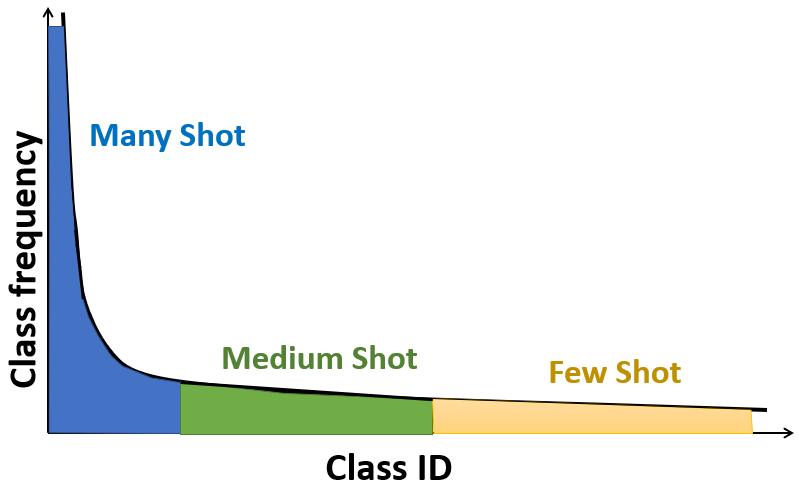}  
  \caption{Long-tail Distribution}
  \label{fig:1}
\end{subfigure}

\caption{Denotes the typical P(y) distributions in the real world data}
\label{tails}
\end{figure}
Regards to the scorer corresponding to class y, $f_y(x)$, we introduce more notations: i.e., $f_y(x) = w_y^T \Phi(x) +b_y$ where $ w_y \in R^K$ are the weights of the last fully connected classification layer correspond to class y. $b_y \in R $ is the bias. Learned embedding corresponds to input x is $\Phi(x) \in R^K$. $K$ is the length of the learnt feature embeddings.  All the weights corresponding to the model are denoted by $W$.
Ideally, label distribution $P(y)$ is uniform. But in real-world applications, $P(y)$ may not necessarily be uniform. It can take the form of a step distribution, or long-tail distribution as indicated in Figure~\ref{tails}. Step distribution is a piecewise constant function with a set of frequent/majority classes and rare/minority classes. In most of the natural data distributions,  the degree of long-tailed ness may vary depending on the application.  These class imbalances in the training data distributions can adversely affect the learnt classifier, causing a bias towards the majority classes when the frequency gap between the majority and minority classes is significant(> 1:10). 
The main goal in a classification problem with skewed label distribution is to avoid misclassifying the rare class instances due to their label scarcity~\cite{survey14,places}.  To this 
end, balanced error is minimized ~\cite{menon2013statistical} in place of the standard error to hypothesise a uniform distribution. 
Balanced error is obtained by averaging the per-class error rates.
  \begin{equation}
  \mathbb{R}_{\text{bal}}(f)=\frac{1}{L}\sum_{y \in [L]} P_{x|y}(y \notin \text{argmax}_{y'\in L} f_{y'}(x))
\end{equation}
  
Therefore, balanced probability can be denoted by,
$P_{bal}(y|x) \propto \frac{1}{L}. P(x|y)$. In~\cite{menon2020long}, authors state that to achieve the goal of minimising the balance error it is necessary to find the best possible scorer for this model: i.e., $f^* \in \text{argmin}_{x \to R^L} \mathbb{R}_{bal}(f)$ Considering Theorem 1 in \cite{collell2016reviving} it can be denoted that,

\begin{equation}
\text{argmax}_{y \in [L]} {f_y}^*(x) = \text{argmax}_{y \in [L]} P_{bal}(y|x)= \text{argmax}_{y \in [L]} P(x|y)
\end{equation}

The theorem states that argmax solution for the best possible scorer is equivalent to the argmax solution of the balanced class probability. The theorem also states that when class conditionals $P(x|y)$ are fixed, changing the class prior $P(y)$ does not affect the optimal scores which ensures that the balanced error remains robust to the class imbalance in the label distribution.  

We use the above set of notations throughout the next sections to elaborate on the different algorithmic-level solutions addressing the problem of learning under class imbalance.

\subsection{Performance Metrics}

Conventionally trained deep models and linear classifiers will favour the majority class significantly given the wider margins 
associated with it leading to many false negatives in the minority class.  For example, in Figure ~\ref{fig:logistic} we showed an example from logistic regression that demonstrates the margin bias that occurred due to the class imbalance.  In such settings, the overall accuracy can still remain at a higher level due to the higher number of train/test instances associated with the majority class. Therefore, standard accuracy can not be considered a fair measure in this scenario as it does not sufficiently reflect the measure of correctly classified instances in different classes. In this section, we summarize the appropriate metrics and techniques for evaluating the models trained on LT data.

\subsubsection{Standard metrics}
We first describe model performance metrics that evaluate the long-tail performance more effectively compared to standard accuracy. \\ \vspace{-3mm}

\noindent{\textbf{Balanced Accuracy}: To resolve the issue with the head class bias, the balanced error is minimized as we described in Section~\ref{problem} to increase the balanced accuracy during long-tail classification. The definition of balanced accuracy is the mean of the sum of the per-class accuracy of all classes as denoted in Equation \ref{bac}. 

\begin{equation}
\text{Balanced Accuracy}=\sum \frac{\text{Per class accuracy}}{\text{Number of classes}} 
\label{bac}
\end{equation}

\noindent{\textbf{Harmonic Mean Accuracy}: Du et al.~\cite{cvpr1} use harmonic mean in computing the balanced accuracy in place of the arithmetic mean. This balanced accuracy is considered important as the harmonic mean is sensitive to the smallest value.  In other words, it reflects the magnitude of the least-performing class accuracy. Therefore, increasing the overall harmonic mean is directly associated with increasing the accuracy of the least-performing minority classes.} \\ \vspace{-2mm}

\noindent{\textbf{Relative Accuracy:} Balanced accuracy does not provide information on the contribution of a certain long-tail method to improve accuracy via alleviating the class imbalance effect. The balanced accuracy can be improved by efficient model selection and training independent of the training distribution. Therefore, it is important that we compare the balanced accuracy of a model that is trained on a balanced training set against a model that is trained on a long-tail distributed training set so that it can convey the actual contribution of the model to alleviate the class imbalance effect. To this end, Zhang et al.~\cite{s4} propose a new metric termed \textit{relative accuracy}. The definition of the metric is as follows.
Let $A_u$ - Empirical upper reference accuracy where $A_u=\max(A_v,A_b)$,
$A_v$ - Vanilla accuracy of the backbone trained on a balanced training set with CE,
$A_b$ - Balanced accuracy of the model trained on a balanced training set with the corresponding long-tail method and
$A_t$ - Balanced accuracy of the model trained on a long-tail train set with the corresponding long-tail method. Then the relative accuracy is given by:-
\begin{equation} \label{eq1}
\text{Relative Accuracy} = \frac{A_t}{A_u}
\end{equation}
Here $A_u$ refers to the maximum of vanilla accuracy of the standard model trained with cross entropy loss with balanced training data and a model trained with the corresponding long-tail method with the same set of balanced training data. Once the empirical upper reference is found we can get the ratio of the empirical upper reference accuracy with the balanced data and the log-tail model accuracy trained with long-tailed data. This would make a true contribution to the proposed long-tail method in alleviating the class imbalance. 
\begin{table}[t]
\begin{tabular}{cccccccccc}
\toprule
Method & \Longunderstack{Focal\\~\cite{lin2017focal}} & \Longunderstack{LDAM\\~\cite{cao2019learning}} &\Longunderstack{ESQL\\~\cite{zhong2019unequal}}& \Longunderstack{BALM\\~\cite{ren2020balanced} }&\Longunderstack{LADE\\~\cite{hong2021disentangling}} &\Longunderstack{Two Stage\\ ~\cite{kang2019decoupling}} &\Longunderstack{BBN\\~\cite{zhou2020bbn}}&\Longunderstack{RIDE\\~\cite{wang2020long}} & \Longunderstack{SADE\\~\cite{zhang2021self}}\\ \hline
ACC&45.8&51.1&47.3&50.8&51.5&49.3 &41.2&55.5 &57.3\\ \hline
RA&79.9&89.2&82.5&88.7&89.1 &86.0&71.9&92.2&92.6\\ \hline
\end{tabular}
\caption{Zhang et al.~\cite{s4} reports results on ImageNet-Lt in terms of accuracy (Acc) and relative accuracy (RA) under 90  training epochs.}
\label{ra}
\vspace{-10mm}
\end{table}

Table~\ref{ra} shows the Accuracy and Relative Accuracy measurements for ImageNet\_Lt for several methods we discuss in later sections. According to the table, ensemble methods such as RIDE and SADE have the highest long-tail classification capability as they are reported to have the highest standard accuracy values compared to other LT classification methods. Simultaneously, these methods have the highest relative accuracy. This indicates the capability of the ensemble methods to alleviate the class imbalance effect, which is 92.2\% and 92.6\%. In other words, the unhandled accuracy drop due to the class imbalance effect is approximately $8\%$ from the class-balanced case. For the other methods, it remains greater than $8\%$. The second highest performance is demonstrated in loss modification methods such as LDAM, Balanced Softmax and Lade, which also have significantly high standard accuracy and relative accuracy. According to the RA metric, Loss modification methods appear to alleviate the class imbalance effect from approximately $9 \%$ more compared to Focal Loss and around $2\%$ more than the two-stage decoupled learning method.
 } \\ \vspace{-2mm}
 
\noindent{\textbf{Per Class Variance: }The prediction stability can be measured by the prediction consistency between two augmented inputs from the same instance, where the prediction consistency can be obtained by the cosine similarity between the learnt embedding of the two augmented inputs. Another approach is to compute the per-class variance of the learnt embeddings for all the trained samples and compare it between head and tail classes. Experimental data showcase that the tail classes tend to have larger feature deviation (high variance)~\cite{jitkrittum2022elm} compared to head classes. This measurement provides an understanding of adjusting the learnt feature distribution accordingly. Similarly, the per-class variance of the learnt embeddings for all the trained and validation instances can be compared. Experiments showcase that tail classes tend to have a higher variance (higher feature deviation) in the validation set than in the training set due to overfitting~\cite{ye2020identifying}. This phenomenon can assist in identifying the tail classes which require additional regularization.} 

\begin{figure}
\centering
\includegraphics[width=0.8\linewidth]{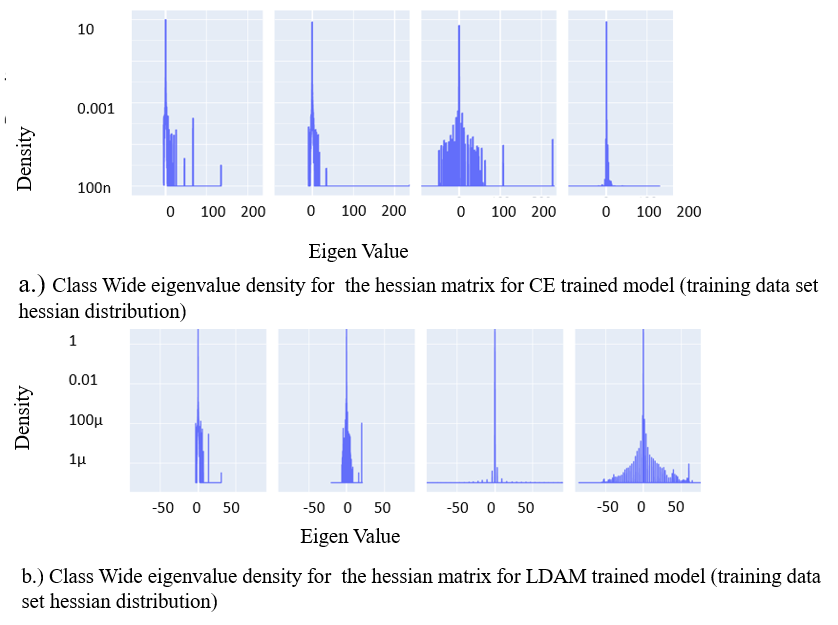}  
 \caption{The figures indicate the class-wise hessian analysis of the loss for CIFAR 10 (IF=100:1) for CE and LDAM methods. The positive concentration of the eigenvalues for the loss function of the head class (0,1) indicates the convergence to the local minima. In contrast, the significant negative curvature for the tail classes(8,9) indicates the convergence of the convergence to saddle point. }
 \label{hess5}
\end{figure}
\subsubsection{Convergence Studies}
 Beyond standard metrics such as balanced accuracy and variants of it, there are some indirect means to measure the performance of long-tail classification and explain the resulting accuracy. Convergence studies involve analysing the loss landscape (i.e., shape, convergence rate).
Rangwani et al.~\cite{rangwani2022escaping} propose an improved Hessian-based analysis of the average loss to analyse the sub-optimal generalisation on tail classes. The proposed framework computes the Hessian of the loss component for each class. Next, the loss landscape is visualised by computing the Hessian Eigenvalue Density~\cite{ghor} for head and tail classes using the Lanczos algorithm. Here, the key focus is on $\lambda _{max}$, the maximum Eigenvalue, and $\lambda_{min}$, the minimum eigenvalue, which indicates positive and negative curvature corresponding to the per-class loss component.  Rangwani et al.~\cite{rangwani2022escaping} claims through the class-wise analysis of loss landscape that if the class-wise Eigenvalue density is concentrated on the positive side, that would indicate convergence of a local minimum. It is also experimentally shown in the same work that for head (majority) classes, there is no significant negative curvature present where Eigenvalue density is concentrated on the positive side which indicates convergence to local minima. On the contrary, in tail classes, the negative curvature is significantly present, which is an indication of converging to a saddle point rather than a local minima.  Saddle points are stable points where the total loss function has a relative minimum in one direction, and a relative maximum in another direction. The authors further demonstrated that when it comes to long-tail classification the objective converges to saddle points where the head classes tend to be in a local minimum and tail classes are prone to converge to a relative maximum. This would result in a significant accuracy drop in the tail classes.  Figure~\ref{hess5}a and  Figure~\ref{hess5}b  show example Eigenvalue density distributions for Cross Entropy loss minimization and LDAM~\cite{cao2019learning} loss, which will be discussed later in this paper under logit adjusted losses. In both cases, Eigenvalue density is concentrated on the positive side for the majority classes 0,1 and considerable negative concentration for rare classes 8,9 of CIFAR 10 dataset. Accordingly, the authors concluded that the accuracy drop in the tail classes can be attributed to the convergence to saddle points.Overall, this Hessian analysis framework can be used to analyze the goodness of long-tail classification techniques as it can investigate convergence properties for the tail classes. The analysis could also be used to guide training parameters (i.e., learning rates, momentum)  selection for long-tail frameworks. 

\subsubsection {Classifier Analysis}
Generally, classifier learning is conducted over the learnt feature distribution by selecting a suitable loss. In some frameworks, classifier learning and feature learning are conducted jointly, while in others, they are conducted in sequential stages. Nevertheless, it is important to evaluate the goodness of the learnt features and the learnt classifier separately to address the issues in each task. We next discuss such analysis techniques that could also be used as a proxy measure of performance in  classification.\\ \vspace{-3mm}

\noindent{\textbf{Class Margin Analysis:}
An ideal classifier must exhibit significant class separability among all its classes. Class separability can be qualitatively studied via inter-class margins. Earlier in Figure~\ref{fig:logistic}, we showed a toy example of class margins under  classification.  The minimum distance from the decision boundary to the class boundary is denoted as the class margin. Margins for classes 0 and 1 are c0 and c1, respectively. Therefore, we can denote the relative margin between the two classes as c0+c1.  When the relative margin is larger, the confusion between the two classes will be reduced. Further, increasing the per-class margin for the tail classes explicitly provides a sufficient generalization for those classes, therefore contributing to reducing overfitting. To this end, Menon et al.~\cite{jitkrittum2022elm} plot the distribution of logit margins between class y and the other remaining classes for all the training instances. Where the class margin $\gamma$ is denoted by Equation ~\ref{margine}. Concerning tail classes, margin distribution with low variance and larger positive mean indicates a robust large positive margin for the tail class, which is followed by higher class separability. In Figure~\ref{margin}, we show the histograms for logit margins for the most and least frequent classes in the CIFAR 100 dataset, respectively. Figure~\ref{margin}a. shows that the class margins are concentrated on the positive side but in Figure~\ref{margin}b., we can see that the class margins significantly lie along the negative side of the x-axis, which indicates that the rare class is inclined to have a higher number of false negatives. Therefore, class margins analysis can be used as a secondary performance metric for long-tail classification. 

\begin{equation}
 \gamma(x,y)=f_y(x)-\max_{y'\neq y} f_{y'}(x)
\label{margine}
\end{equation}

\begin{figure}[t]
\vspace{-20mm}
\includegraphics[width=\textwidth]{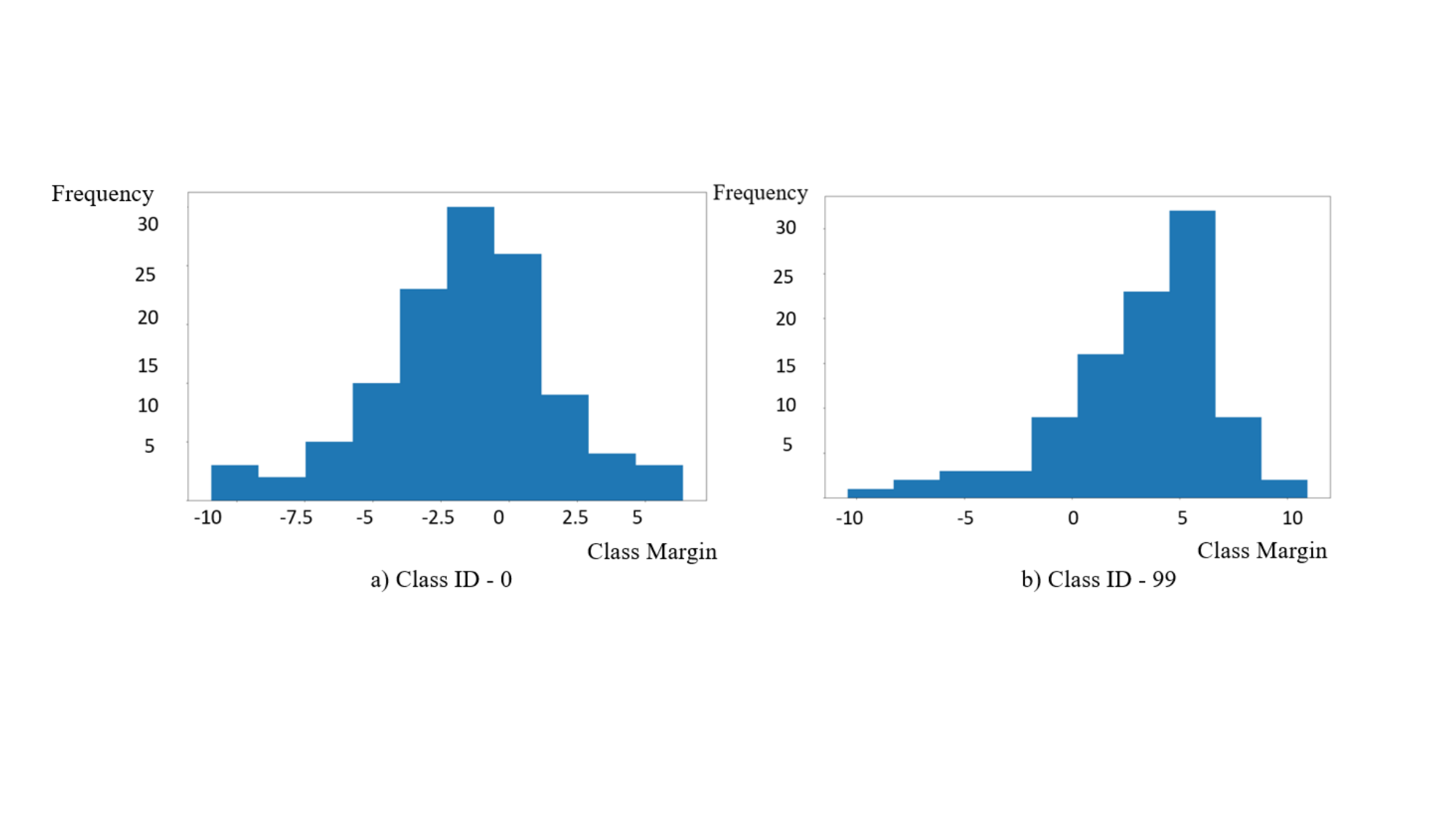}  
\vspace{-30mm}
\caption{Logit margin distribution of the most frequent (class id-0) and the least frequent (class id-99) class for CIFAR 100 (imbalance factor  =100:1) under CE-based model training with no long-tail adjustments. }
\label{margin}
\end{figure}
\subsubsection{Feature Distribution Analysis}

Similar to classifier margin analysis, latent feature distribution can also be used as a proxy measurement to evaluate the performance of various long-tail classification methods. This can be done in three aspects: feature space balancedness, feature compactness, and tail class overfitting. \\ \vspace{-3mm}

\noindent{\textbf{Feature Space Balancedness:} }
In~\cite{balanced} the authors showed that in contrast to supervised cross entropy loss, unsupervised contrastive learning is robust to the training data distribution imbalances. The authors argued that this phenomenon is caused by the relatively balanced feature space generated by unsupervised contrastive learning methods for all the classes, and therefore, the balancedness of the learnt feature space can be used as a metric to compare different algorithms that can handle the imbalances in the training data. To measure this, the authors introduce a metric to evaluate the balancedness of the learnt feature space based on the intuition that leant representations denoted by $v_i$ from all the available classes must have a similar degree of linear separability. In practice, the degree of the linear separability of the learnt features is computed based on the accuracy of the linear classifier trained on them. Inspired by this, authors propose the balancedness metric for feature space V is denoted in Equation~\ref{eq2}.

\begin{equation}
\beta(V)=\frac{1}{C}\sum_{i,j}^C \exp(\frac{-|a_i-a_j|^2}{\sigma})
\label{eq2}
\end{equation}

For a linear classifier (W,b), the classification accuracy over C classes is represented by $ a_1,a_2,...,a_C$.
$\sigma$ is a constant scaling parameter that is dependent on the dataset. When the class-wise accuracy is equal, the metric reaches its maximum possible value of 1. } \\ \vspace{-2mm}

\noindent{\textbf{Feature Compactness}:
 In long-tail classification learnt embeddings of the tail classes tends to deviate from its class centre occupying a large space in the feature space. This is causing errors in generating accurate class boundaries for those classes, leading to many false negatives. Feature compactness can be qualitatively analyzed through TSNE or UMAP plots in 2D. The feature distribution for the validation set of the head (many shot) vs tail (few shot) classes can be visualized to compare the cluster compactness~\cite{de2023long}. It also can be quantitively analyzed through cluster compactness measures or variance measures. The long-tail classification frameworks can be improved to implicitly address the feature deviation issue and establish a reasonable compactness for tail classes which is equivalent to the head classes. The idea behind the compactness analysis is to improve the per-class accuracy from the perspective of the class embeddings.} \\ \vspace{-2mm}

\noindent{\textbf{Tail Class Overfitting:}.
When there is a significant drop from training accuracy to test accuracy for tail classes,  it is crucial to have a strategy to verify whether this phenomenon is caused by overfitting or vice versa. 
Ye et al.~\cite{ye2020identifying} empirically show that features $f(\theta)$ (last layer embeddings) of the training and test data deviate from each other, and that is an indication of the degree over-fitting. This phenomenon can also be visualised using TSNE or UMAP plots in 2D. This analysis also can be used to explain the accuracy of drop-in tail classes. This is also useful to decide the optimal level of oversampling or scaling up (reweighting), which must be conducted on the tail classes to see a significant accuracy enhancement with a sufficient generalization or it has reached the limit. As in the work of Ye et al.~\cite{ye2020identifying}, TSNE or any other similar visualization of the features $f(\theta)$ of training and test set can be used to understand the extent of the deviation. If this deviation is significant, then it implies that a linear classifier learned to classify training instances does not successfully classify test instances as it does not occupy the same feature space. The other key observation is that when the deviation between the training and the test features is larger, the score/logits/decision values corresponding to the test instances grow smaller. Therefore, we can use this feature deviation analysis to select the classes that require maintaining the class score of the training instances at a higher value so that despite the accuracy drop for test instances due to the feature deviation, it can still belong to the correct side in the classifier. }\\ \vspace{-2mm}

\subsection{Datasets}

We next describe the most commonly used datasets and experiment settings in deep classification works. Typically, the classes are split into three subsets depending on their cardinality many shots, medium shots and few shots. The cardinality thresholds for the division are usually 100 and 20. There are metrics such as class Imbalance Factor (i.e., IF - the ratio between the most frequent and least frequent class), standard deviation and mean of class frequency distribution, and the Gini Coefficient that are used to estimate the long-tailness of the datasets. For instance, Yang et al.~\cite{s1} use Gini coefficient to compare the long-tailed ness of different datasets. Gini coefficient is interpreted using the cumulative share of class frequency plotted against the cumulative share of classes from lowest to highest frequency. When the dataset is balanced Gini coefficient reaches zero. We provide a summary of the commonly used datasets in experiment settings in  classification in Table~\ref{tab1}.  \\ \vspace{-2mm}

\noindent{\textbf{CIFAR10-LT and CIFAR100-LT}: The long-tailed datasets CIFAR10-LT and CIFAR100-LT are curated from original CIFAR datasets.  CIFAR-100-LT consists of 100 classes, 500 training instances in the most frequent class and 500/IF number of instances in the least frequent class. CIFAR-10 includes 10 classes with 5,000 training images in the head class and 5,000/IF images in the least frequent class. Class imbalance ratios of 50, 100, and 200 are commonly used in long-tail classification experiments for these datasets. The test sets are balanced datasets containing 1,000 images and 100 images for  CIFAR10-LT and CIFAR100-LT, respectively.} \\ \vspace{-2mm}

\noindent{\textbf{ImagNet-LT}: Originally, ImageNet is a uniformly distributed dataset containing 1,000 classes where each class consist of 1,300 images. The ImageNet\_Lt, a long-tailed version of the dataset, is generated via Pareto distribution by allocating 1,280 images for the majority class and 5  images to the rarest class. The validation dataset remains uniformly distributed. The dataset's accuracy is estimated by dividing it into three main categories, many shot, medium shot, and few shot. The cardinality threshold for the division is 100 and 20.} \\ \vspace{-2mm}

\noindent{\textbf{iNaturalist}~\cite{iNaturalist18}}: The 2018 version of the iNaturalist dataset is used as a state-of-the-art dataset for long-tail experiments. This dataset is inherently imbalanced and follows a long-tail distribution. This collection of shared images was taken by naturalists worldwide observing biodiversity and nature. The 2018 version consists of 8,142 categories (classes) and 437,513 images. Typically, the imbalance ratio used in the dataset is up to 500.}  \\ \vspace{-2mm}

\noindent{\textbf{LVIS~\cite{gupta2019lvis}}: Large Vocabulary Instance Segmentation is a long-tailed dataset that consists of 1,230 classes where 454 classes are considered rare classes, 461 classes are categorized as common, and 315 are considered frequent classes. The cardinality of the class thresholds is 100 and 10 for many shots, medium shots, and few shots. The class imbalance ratio used for the experiments is up to 26,000. This is the most challenging dataset among the other SOTA long-tail datasets due to its extreme imbalance and longer tail.} \\ \vspace{-2mm}

\noindent{\textbf{Places-LT~\cite{places}}: This is derived from Places-2 dataset to follow a Pareto Distribution. The training set consists of 62,500 images from 365 classes. The most frequent class has 4,980 images per class, and the cardinality of the least frequent class is 5. The validation and test sets are uniformly distributed, the former having a cardinality of 20 images per class and the latter 100 images per class. The used class imbalance ratio is up to 996.}  \\ \vspace{-2mm}

\begin{table}[t]
\centering
\begin{tabular}{ccccc}
\toprule
\textbf{Dataset} & \textbf{Maximum Class Imbalance Factor} & \textbf{Number of classes} & \textbf{Head class frequency}\\\hline\hline
 CIFAR10\_LT  &100-200 &10& 5,000\\\hline
CIFAR100\_LT & 100-200& 100 & 500 \\\hline
 ImageNet\_Lt & 256 & 1,000 & 1,280  \\ \hline
 iNaturalist & 500 &8,142 &N/A\\ \hline
 LVIS & 26,148 & 1,230 & 26,148 \\ \hline
 Places\_LT & 996 &365 & 4,980  \\ \hline
\end{tabular}
\caption{Summary of the long-tail datasets' specifications. N/A-information is not available in literature }
\label{tab1}
\vspace{-10mm}
\end{table}

\section{Taxonomy}

\begin{figure}
    \centering
  \includegraphics[width=1\linewidth]{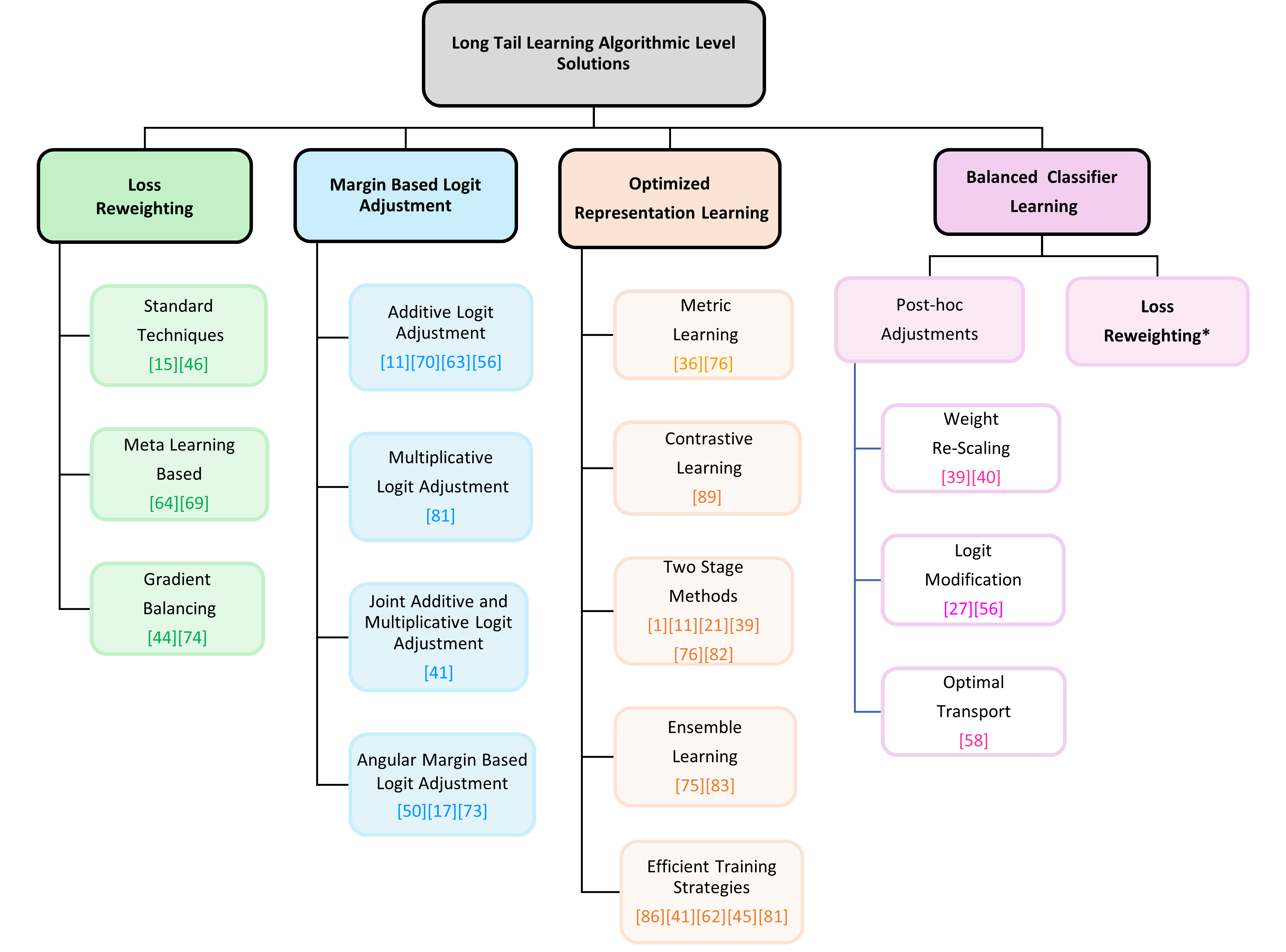} 
    \caption{High-level categorization of algorithmic solutions for deep  classification. Loss Rewighting* branch appears under Balanced Classifier Learning is redirected to Loss Reweighting main branch }
    \label{tax}
\end{figure}

Class imbalance handling for long-tail settings is conducted at both the data and algorithmic levels.
Data-level techniques such as over-sampling, under-sampling, square root sampling and progressively balanced sampling~\cite{126,127,128,129} are commonly used to resolve the class imbalance effect by mimicking a balanced class distribution. In addition to these techniques data augmentation~\cite{150,151}  is also successfully used to alleviate class imbalance issues. Augmentations to the rare class instances enhance the robustness and the per-class accuracy of those classes. Even though data-level solutions add a significant level of accuracy to long-tail settings, they alone can not eliminate the class imbalance effect. This is mainly due to the overfitting and underfitting occurring with the various sampling methods. Secondly, most data-level methods increase the tail class accuracy at the cost of the head class accuracy. i.e., undersampling would increase the underfitting there by the bias of the head classes. Therefore, algorithmic level solutions such as loss modifications and improvements to the model architecture came into play as they can work complementary to the data level solutions. They are also highly important due to their ability to enhance the long-tail classification accuracy without altering the original training data distribution. Conventionally data level and algorithmic level solutions are used in a hybrid framework to operate complementary to each other to gain better model performance. Throughout this survey, we mainly focus on the algorithmic-level solutions. We categorise existing algorithmic level approaches into four broad categories: \emph{i) Loss reweighting, ii) Margin Based Logit Adjustment, iii) Optimized Representation Learning} and \emph{iv) Balanced Classifier Learning} as illustrated in Figure~\ref{tax}.

\subsection{Loss reweighting}
\begin{equation}
l_{CE}(y,f(x)) = -\frac{1}{w_x} \text{log} \frac{e^{f_y(x)}}{\sum_{{y’} \in Y} e^{f_{y’}(x)}}
\label{rewe1}
\end{equation}
reweighting methods apply weights based on different training samples, based on the class frequency or classification hardness. In Equation~\ref{rewe1}, $w_x$ denotes a static or dynamic weighting factor on the loss component during optimization. These techniques try to enhance the influence of the tail classes. However, this may impair the classification accuracy of the head classes and overfit the rare classes~\cite{cui2019class}. Therefore, it is important to add reweighting optimally. Overall, reweighting methods demonstrate a notable improvement in accuracy compared to conventional cross entropy (CE) approaches.
  We further categorize reweighting methods into 1) Standard techniques, 2) Gradient Balancing techniques, and 3) Meta-learning-based approaches. We discuss these methods in detail in Section 4.
\subsection{Margin Based Logit Adjustment}
\begin{figure}[bt]
\includegraphics[width=0.6\linewidth]{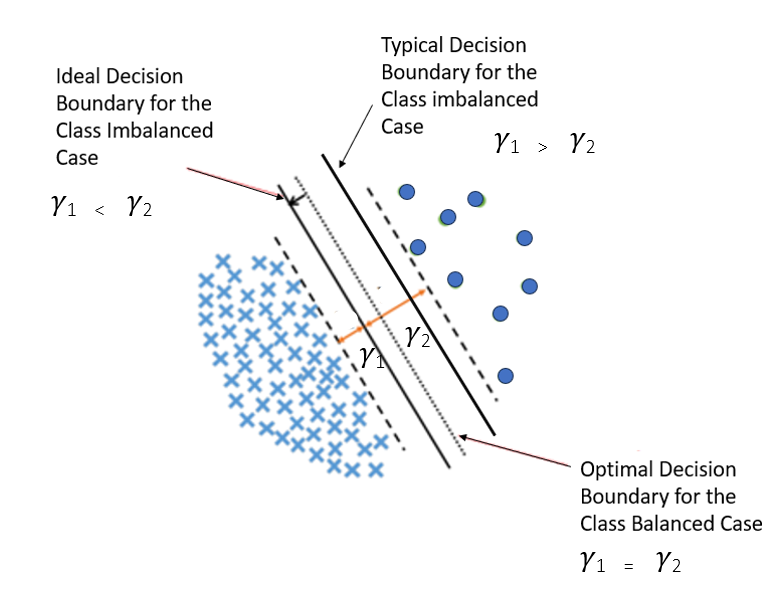}  
  \caption{Optimal decision boundary for class imbalanced case - A larger margin is enforced on the rare class and compared to the frequent class. Here, $\gamma $ denotes the per-class margin.}
  \label{margin1}
\end{figure}
Margin-based logit adjustment methods adapt classifiers to the label distribution shift. i.e.,  training distribution to the uniform test distribution. Under these methods, model score/logit values are modified based on parameters derived using the class margin values, i.e., addition adjustments or/and multiplicative adjustments, thereby modifying the SoftMax cross entropy computation.  As shown in Figure~\ref{margin1} these method encourages the classifier to have optimal decision boundaries so that enforcing a large relative margin between a tail and a head class by regularizing the tail class more strongly compared to the head class without reducing the capability to model the head class. To this end, it requires a data-dependent or label-dependent regularizer rather than a standard L2 regularizer which only depends on the model parameters. Therefore, the most conveniently interpretable data-dependent property, "margins of the training samples", is used for regularization purposes. According to Bartlett et al. ~\cite{baret} and Wei et al.~\cite{wei}, encouraging a larger margin can be indicated as a means of regularization because the inverse of the minimum margin among all the training instances correlates with the generalization error bounds.  Section 5 describes intuition for logit adjustment from the fundamentals of Support Vector Machines, and subsequently, different deep learning-based logit adjustment techniques such as Additive Logit Adjustment, Multiplicative Logit Adjustment, Joint Additive and Multiplicative Adjustment, and Angular Logit Adjustment are discussed.
\subsection{Optimized Representation Learning}
\begin{figure}
\centering
\includegraphics[width=1\linewidth]{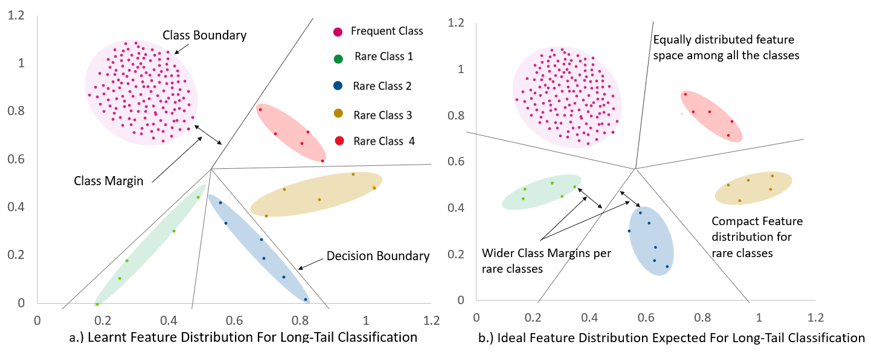}
\caption{ (a) This figure denotes the distribution of the penultimate layer embeddings in a 2D space in the class imbalanced 
case for linearly separable data. Important factors to note are 1) feature distribution in the hyperspace, 2) class margins, and 3) feature diffusion in each 
class. (b) Shows the expected optimal feature distribution under class imbalance. Symmetrically distributed class 
centres in the hyperspace, higher feature compactness, clear class separability and generalisation of the class margins that are obtained with the representation learning
}
\label{cc1}
\end{figure}

Loss reweighting and Margin-based logit adjustment tend to enhance the rare class performance significantly but they sacrifice the accuracy of frequent classes. The above methods do not control the learned feature distributions explicitly~\cite{jitkrittum2022elm}. This may lower the compactness of the learnt embeddings of the tail classes. Then, the intra-class dimensions get larger, resulting in the diffusion of the embeddings~\cite {ye2020identifying}.
This phenomenon will increase the misclassification error in the tail classes. These key characteristics of the learnt representation in a long-tail setting are illustrated as follows. consider a simple 2D learnt feature distribution with linearly separable data as indicated in Figure~\ref{cc1}a, where frequent/head class occupying a vast space in the feature space have broader margins compared to the tail classes and possess higher feature compactness. Tail class embedding tends to diffuse in a wider spacing, resulting in lower-class compactness.  Ideally, optimized (class imbalance handled) learnt feature representation should share the feature space evenly among different classes, enhancing the balancedness, higher feature compactness for all the classes and sufficient feature separability (larger margins and strong regularization) among all classes as indicated in Figure~\ref{cc1}b.

 To address the issues in the learnt representation, two-stage learning approaches and multi-branch networks are employed to handle both representation learning and classifier learning optimally. In sequential multi-stage learning tasks, the first stage focuses on learning the representation using the original data distribution, while the second stage is dedicated to classifier learning. Multi-branch networks typically have parallel branches to learn representation and the classifier.  With these methods, it is convenient to use techniques to optimize the learned feature distribution independently to avoid the diffusion of the rare classes. In Section 6, we discuss metric learning and contrastive learning approaches used to improve the learnt feature distribution. Subsequently, we also discuss two-stage learning, ensemble learning techniques, and  Efficient Training Strategies used to enhance representation as per Figure~\ref{cc1}b.

\subsection{Balanced Classifier Learning}

Balanced classifier learning is conducted to avoid bias towards majority classes as it contains a larger portion of the training data. It is important to decide the level of classifier balancing depending on the application and the class imbalance ratio. i.e., In fraud detection, the fraud class should be identified with higher accuracy given the low level of training associated with the fraud class. Therefore, this classifier needs an additional balancing towards the minority fraud class. 
 Class balancing is conducted by exploiting a balanced distribution for fine-tuning the model (data-level solution) or applying a re-balancing technique (algorithmic level) specifically to the classifier. Other than that post hoc calibrations to the model predictions or domain shift calibrations can also be used to obtain the necessary balance. However, experimental evidence suggests that these approaches may not be sufficient to fix the imbalance problem, particularly in scenarios with an extreme class imbalance or many classes. Classifier balancing can complement the reweighting and representation learning methods~\cite{kang2019decoupling, alshammari2022long}. Algorithmic-level balanced classifier learning techniques are further classified as Loss reweighting and  Post-hoc techniques. We describe these methods in Section 7.

\section{Loss reweighting} 
\label{rew}

This section discusses the reweighting methods according to the division shown in Figure~\ref{tax} Loss Reweighting branch. Firstly, we introduce the standard static reweighting methods based on the frequency or classification hardness. Secondly, the loss gradient-based reweighting techniques are discussed. Thirdly, dynamic reweighting based on meta-learning approaches is presented.  One main consideration to note is carefully selecting the corresponding regularization technique when using the reweighting method. i.e., Byrd et al.~\cite{byrd2019effect} demonstrated that importance reweighting without regularization does not introduce any performance gain. 

\subsection{Standard Techniques}
Typically, a class-balanced loss assigns sample weights
inversely proportionally to the class frequency. This simple
heuristic method has been widely adopted. However, recent work on training from large-scale, real-world,
long-tailed datasets reveals poor performance when
using this strategy. Instead, they use a ``smoothed'' version
of weights empirically set to be inversely proportional to the square root of class frequency.  In~\cite{cui2019class}  
a novel theoretical framework is proposed to characterize data overlap
and calculate the effective number of samples in a model and loss-agnostic manner. A class-balanced reweighting term, that is inversely proportional to the effective number of
samples is added to the loss function.  The class-balanced
softmax cross entropy loss is given by Equation~\ref{c1}. where $\beta=(N-1)/N$ and N is the volume of the set of all possible data in the feature space of this class. To elaborate on the intuition behind N,  it can be denoted as in Equation~\ref{beta}. \\ \\
\noindent\begin{minipage}{.5\linewidth}
\begin{equation}
l_{CE}(y,f(x))=-\frac{1-\beta}{1-\beta^{n_y}}\log{\frac{e^{f_y(x)}}{\sum_{j=1}^{C}e^{f_j(x)}}}
\label{c1}
\end{equation}
\end{minipage}%
\begin{minipage}{.5\linewidth}
\begin{equation}
N=\lim_{n \rightarrow  \infty} \sum_{j=1}^n \beta^{j-1} =\frac{1}{1-\beta}
\label{beta}
\end{equation}
\end{minipage}\\ \\
$n_y$ is the number of samples corresponding to class $y$. Even though this method balances the weight of different classes, it has a limited capacity to prioritise between easy and hard samples. Therefore, the focal loss~\cite{lin2017focal} denoted in Equation~\ref{focal1}  comes into play to emphasise more weight on the hard samples.
When compared to softmax probability, sigmoid has some special features. It does not assume mutual exclusiveness; therefore, it is compatible with real-world scenarios where some of the classes are very similar to each other. Further, as sigmoid has an individual predictor for each class, the prediction can be considered independent and can unify the single-label classification with muti-label classification when real-world examples may have multiple semantic labels. 

\begin{minipage}{0.5\textwidth}
 \begin{equation}
l_{focal}(y,f(x))=\sum_{i=1}^C \frac{e^{-f_i^t(x)}}{1+e^{-f_i^t(x)}}[log(\frac{1}{1+e^{-f_i^t(x)}})]
\label{focal1}
\end{equation}

\end{minipage}
\begin{minipage}{0.5\textwidth}
   
\[
where \hspace{10mm}\\
    f_i^t(x) =
\begin{cases}
    f_i(x),& \text{if } i==y\\
    -f_i(x),              otherwise
\end{cases}
\] 
\end{minipage}

Focal loss  can be integrated with class balanced component 
\begin{equation}
l_{weighted focal}(y,f(x))= -\frac{1-\beta}{1-\beta^{n_y}} \sum_{i=1}^C \frac{e^{-f_i^t(x)}}{1+e^{-f_i^t(x)}}[log(\frac{1}{1+e^{-f_i^t(x)}})]
\end{equation}

\subsection{Meta Learning Based Approaches}

Ren et al.~\cite{400} demonstrate an approach to enhance the effectiveness of the training objective through a weighted loss in place of the average loss combined with the instantiating of meta-learning. Here validation loss is being used as the meta objective and the algorithm takes one gradient descent step over each iteration. This method claims to have no additional hyper-parameters to learn and also the expensive optimization tasks are simplified by an online approximation. The weighted objective discussed in this work can be denoted as follows,
\begin {equation}
W^*(\omega) = \text{arg min}_W {\sum_{i=1}}^N \omega_i l(y_i,f(x_i,W))
\end{equation}
The optimal selection of the w values on each loss component of the training examples is conducted based on the validation performance of the model. Let $V=(x_i^v,y_i^v)_{i=1}^M $ where $M << N$. Where $V$ is a clean and unbiased validation set used to optimize the parameters $\omega$.

\begin{equation}
    \omega^*=\arg \min_{\omega,\omega\geq 0} \frac{1}{M} \sum_{i=1}^M {l}^v(y_i^v,f(x_i^v,W^*(\omega)))
\end{equation}
There is another approach proposed in \cite{shu2019meta} to adaptively learn sample weights. The weighting function is formulated as a multi-layer perceptron network with one hidden layer termed a Meta-Weight-Net and exploiting a meta-data (relatively small unbiased validation set) to train the parameters of the network. This network contains 100 nodes. All the hidden nodes are associated with ReLu activation function and the output is linked with the sigmoid activation function to ensure that the output values are in the range of [0,1]. Under class imbalanced scenarios, this sample reweighting strategy increases the model robustness by the weights $\nu(l(y,f(x,W)),\theta)$ imposed on the loss of the $i^{th}$ sample. $\nu$ represents the weight net, and the $\theta$ denotes the parameters of the weight net. The optimal classifier parameters $w$ are obtained by minimizing the loss indicated in Equation~\ref{loss12}.

\noindent\begin{minipage}{.6\linewidth}
\begin{equation}
    W^*(\theta)=\frac{1}{N}{\sum_{i=1}}^N \nu(l^{train}(y_i,f(x_i,W)),\theta) . l^{train}(y_i,f(x_i,W)) \label{loss12}
\end{equation}
\end{minipage}%
\begin{minipage}{.4\linewidth}
\begin{equation}
    \theta^*=\arg \min_{\theta,\theta\geq 0} \frac{1}{M} \sum_{i=1}^M {l}^v(y_i^v,f(x_i^v,W^*(\theta)))
\end{equation}
\end{minipage}

However, a major drawback of weighting methods is the challenges encountered in the optimization process as the selection of weights is sensitive to the optimizer that is being used~\cite{huang2019deep}. 
\subsection{Gradient Balancing}
Training instances from the head classes are significantly high compared to the tail classes in a long-tail dataset. These head class instances result in large proportions of negative instances for the tail class. Therefore, the positive and negative sample gradients on a certain tail class become extremely imbalanced, resulting in a biased classifier for the head classes. This phenomenon will reduce the accuracy of the tail classes. To resolve this issue, Li et al.~\cite{li2019gradient} use the harmony of the gradient contribution of different training instances. The authors assign a harmonizing parameter to the gradient of each training instance. This parameter is generated based on the number of training instances with similar characteristics about the gradient density.

Later, Wang et al.~\cite{wang2021seesaw} propose a solution based on the example of a seesaw where two objects are positioned on each end of it. To balance the objects, a feasible solution is to set a short length for the arm where the heavier object is placed.  Similarly, the class balance can be obtained by scaling down the dominating gradients of the negative instances of the minority class. The proposed seesaw loss dynamically balances the positive and negative gradients for each class with two factors termed as mitigation factor $\mathbb{M}$ and compensation factor $\mathbb{C}$ that are complementary to each other. Key features of seesaw loss are: 1) It is a dynamic loss that analyses the cumulative training sample ratio between different classes and instance-wise misclassifications as the training occurs, 2) The loss is self-calibrated as the mitigation, and the compensation factors reduce the increasing false positives for the minority classes, and 3) The seesaw loss is robust to the training data distribution and the nature of the sample. It can gradually approximate the true data distribution as the training occurs and provides the necessary balancing.
For a training sample (x,y), the seesaw loss is given by

\begin{equation}
 l(y,f(x)) =-\log \frac{ e^{f_y(x)}}{\sum_{y’\neq y} \mathbb{S}_{y’y}. e^{f_{y’}(x)}+ e^{f_y(x)}}
\end{equation}

where$\mathbb{S}_{y’y} $is the tunable balancing parameter between different classes $y$ and $y’$  is given by 

$\mathbb{S}_{yy’ } = \mathbb{M}_{yy’} \mathbb{C}_{yy’}$.

\noindent\begin{minipage}{.5\linewidth}
\begin{equation}
    \mathbb{M}_{yy’} =
\begin{cases}
    1 ,& \text{if } N_y \leq N_{y’}\\
    (\frac{N_{y’}}{N_y})^p,              & \text{if } N_y \geq  N_{y’}
\end{cases}
\label{m}
\end{equation}
\end{minipage}%
\begin{minipage}{.5\linewidth}
\begin{equation}
    \mathbb{C}_{yy’} =
\begin{cases}
    1 ,& \text{if} \sigma_{y’} \leq \sigma_y\\
    \frac{\sigma_{y’}}{\sigma_y},              & \text{if} \sigma_{y’} \geq  \sigma_{y’}
\end{cases}
\label{c}
\end{equation}
\end{minipage}

The Explanation for mitigation factor $\mathbb{M}$ is given by Equation~\ref{m}. Where $N_y$ is the cumulative number of training instances processed so far during the training corresponding to the class $y$, p is a tunable hyperparameter to adjust the level of mitigation. Subsequently, elaboration for compensation factor $\mathbb{C}$ is given by Equation~\ref{c}. Where $\sigma_y$ is the predicted softmax probability class y.q is a tunable hyperparameter to control the compensation scale.

\subsection{Summary}

To summarise, loss reweighting methods introduce $2\%-4\%$  balanced accuracy improvement to the standard CE loss considering the works discussed in the survey. These frequency-based reweighting methods conveniently apply to any loss function without any expensive offline training stage. Reweighting can complement margin-based logit adjustment techniques and representation learning techniques to obtain added performance increment. A frequency-based reweighting scheme can improve LDAM and additive logit adjustment discussed in Section~\ref{logi}. Multistage representation optimization discussed in Section~\ref{opt} can also be improved with frequency-based reweighting. However, during reweighting hard samples are usually highly weighted. However, applying larger weights may be unreasonable when the label noise is present. Further, the minority class samples are also highly weighted compared to majority classes but this also may not be fair when the class distinctiveness of the majority class is low. Therefore, the reweighting must be optimized while considering all the other factors, including frequency.  To resolve issues with noisy data meta-learning-based reweighting has come into play. Where a validation set is used to guide the reweighting parameters, during conventional reweighting methods, hard and easy example mining is conducted during the procedure. Overlapping instances could be removed, and the most effective instances may be selected.  Gradient balancing methods are proposed to overcome the discarding of training data. These gradients have also shown significant accuracy enhancement compared to standards cross entropy loss. i.e., Seesawloss indicates $6\%$ accuracy improvement for ImageNet\_Lt.

\section{Margin Based Logit Adjustment}
\label{logi}
\label{logit3}

This section describes the loss modification based on logit adjustment, which encourages a large relative margin between the minority and majority classes. Generally, The extent of the logit adjustment is decided based on the relative class imbalance between two classes. We mainly focus on the logit adjustment techniques applied during training time and the importance of applying them in different training stages. We also provide the intuition for the logit adjustment starting from a simpler case, applying asymmetric logit margins for support vector machines-based linear models. In the later part, we discuss logit adjustment in long-tail classification in the context of deep learning. As indicated in Figure~\ref{tax}, branch margin-based Logit adjustment is divided into four main categories which will be discussed in this section. 
\subsection{Preliminaries-Cost Sensitive Support Vector Machines}
We first describe Support Vector Machine based class imbalance handling techniques as preliminary work since they are associated with class margins and provide key intuition in loss modification techniques related to deep nets, which we will discuss later in this section. In the context of binary classification with balanced linearly separable data, two parallel hyperplanes (class boundary) can be defined to separate the two classes so that the distance between the planes is maximum. In this scenario, the distance between the class boundaries (hyperplanes) is termed the margin, and the maximum margin hyperplane that lies in the middle is the optimal decision boundary for the classes. However, when the data contains outliers and the overfit is higher, a soft margin is implemented by allowing misclassifications in some instances. Soft-margin SVM is defined using a hinge loss as indicated in Equation~\ref{hinge} and used for maximum margin classification for SVM. Optimization over the entire dataset is denoted by Equation ~\ref{obj}.
\begin{equation}
l(y,f(x))=\max(0,1-y_if(x_i))
\label{hinge}
\end{equation}

\begin{equation}
Total Loss = \lambda ||w||^2 + \frac{1}{N} [ \sum_{i=1}^N  \max(0,1-y_if(x_i))]
\label{obj}
\end{equation}
Veropoulos et al.~\cite{veropoulos1999controlling} introduce a cost-sensitive strategy referred to as a different error costs method to handle the class imbalance in real-world binary classification problems. In this approach, the soft margin SVM objective function is redesigned to incorporate the cost of misclassification $C^+$ for positive class samples and the cost of miss-classification $C^-$  for negative class samples when the data is not linearly separable. 

\begin{subequations}\label{lo}
\begin{align}
\text{min} \frac{1}{2} W.W + C^+ \sum_{i|y_i =+1}^N \xi_i + C^{-1}  \sum_{i|y_i =-1}^N \xi_i  \\
\text{s.t }  y_if(x_i) > 1-\xi_i \text{ ; }
_i \geq 0, i=1,...,N 
\end{align}
\end{subequations}

Where  $\xi$ is the parameter to control the soft margin. In class imbalance scenarios by setting the minority class misclassification cost to a larger value compared to the majority class misclassification cost can decrease the effect of the imbalance on the classification margin. When the misclassification cost $C$ is large, the slack variables are minimal, and the optimization problem is reduced into a standard SVM problem, in which case the decision boundary is computed so that there is an equal margin for both classes rather than assigning a larger margin for the minority class.

 \subsection{Loss Modification in Deep Architectures with Logit Adjustment}
\subsubsection{Loss modification by Additive Logit Adjustment:}
Under loss modification, promising results are achieved through incorporating logit margins to the standard softmax cross entropy. The main goal of these techniques is to allocate larger margins for the rare classes compared to the frequent classes to increase the separability of the rare classes. It also alleviates the intrinsic bias of the softmax cross entropy under long-tail scenarios by addressing the distribution shift from training (long-tail) to test (uniform) distribution. Reformulation of Equation~\ref{ce} with added logit margins defines the generic description of the method. Where $\Delta_{yy’} $ can be defined as a margin between the classes $y$ and $y’$ or in other words the gap between the scores of class $y$ and $y’$.

\begin{equation}
 l_{CE}(y,f(x)) = \log[ 1+ \sum_{y’ \neq y} e^{\Delta_{y’y} + f_{y’} (x)- f_y (x)} ]
 \label{ce}
\end{equation}

Cao et al.~\cite{cao2019learning}  introduce an approach that is a multi-class extension of the hinge loss. As non-smoothness in the hing-loss poses optimization difficulties, cross entropy loss denoted with enforced margins is used as a smooth relaxation for the hinge loss as denoted in  Equation~\ref{cao}.
 A per class 
margin into the softmax cross entropy using an additive component on the score value corresponding to the given label. 
$f_{\emph{y}}(\emph{x})\to f_{\emph{y}}(\emph{x})-\delta_{\emph{y}}$

\begin{equation}
\emph{l}(\emph{y},\emph{f}(\emph{x}))=\text{log}[1+\sum_{\emph{y}'\neq \emph{y}}e^{\delta_\emph{y}}.e^{f_{\emph{y}'}(\emph{x})-f_{\emph{y}}(\emph{x})}]
\label{cao}
\end{equation}
where $\delta_\emph{y} \propto P(\emph{y})^{\frac{1}{4}}$ which enforces an additional regularization on rare positive classes. Conventional loss functions such as softmax cross entropy and sigmoid cross entropy which are used in the classification 
tasks have a mitigation effect on the 
non-ground-truth classes. Tan et al.~\cite{tan2020equalization} propose an additive adjustment on the non-ground truth class score values to reduce the
 influence of negative samples for the less frequent classes.
 $f_{\emph{y’}}(\emph{x})\to f_{\emph{y’}}(\emph{x})+\delta_{\emph{y}}$
 
 \begin{equation}
\emph{l}(\emph{y},\emph{f}(\emph{x}))=\text{log}[1+\sum_{\emph{y}’\neq \emph{y}}e^{\delta_{\emph{y}’}}.e^{f_{\emph{y}’}(\emph{x})-f_{\emph{y}}(\emph{x})}]
\end{equation}
where $\delta_{\emph{y}’}$ is a non decreasing transform of $P(\emph{y}’)$.
As discussed above long-tailed scenarios result in an intrinsic bias in the softmax function. Ren et al.~\cite{ren2020balanced} explicitly models the 
label distribution shifts during the test time through additive adjustments to the score functions.
 $f_{\emph{y’}}(\emph{x})\to f_{\emph{y’}}(\emph{x})+\frac{1}{4}\log [n_{\emph{y’}}]$ 
 and $f_{\emph{y}}(\emph{x})\to f_{\emph{y}}(\emph{x})+\frac{1}{4}\text{log } [n_{\emph{y}}]$
\begin{equation}
\emph{l}(\emph{y},\emph{f}(\emph{x}))=\text{log}[1+\sum_{\emph{y}'\neq \emph{y}} (\frac{n_{\emph{y}'}}{n_\emph{y}})^{\frac{1}{4}}.e^{f_{\emph{y}'}(\emph{x})-f_{\emph{y}}(\emph{x})}]
\end{equation}
Ren et al. \cite{ren2020balanced} also provide a theoretical validation for the inherent bias of the Softmax function when dealing with long-tailed scenarios. As a solution, they introduce the Balanced Softmax function based on a probabilistic perspective, aiming to tackle the train-to-test distribution shift. Additionally, the authors demonstrate that optimizing the proposed Balanced Softmax cross entropy loss is equivalent to minimizing the bound on generalization error.
The proposed balanced error can be reformulated in the following form.\\
\begin{equation}
\emph{l}(\emph{y},\emph{f}(\emph{x}))=\text{log}[1+\sum_{\emph{y}'\neq \emph{y}} (\frac{n_{\emph{y}'}}{n_\emph{y}}).e^{f_{\emph{y}'}(\emph{x})-f_{\emph{y}}(\emph{x})}]
\end{equation} 

Note that in this formulation the power coefficient $\frac{1}{4}$ in Equation 15 is replaced with 1. 
Menon et al.~\cite{menon2020long} also propose a similar approach where a tunable hyperparameter is used in place of the power coefficient $\frac{1}{4}$ of the above method. Also note that $\frac{n_{\emph{y}'}}{n_\emph{y}}$ is replaced with $ \frac{\pi_{\emph{y}'}}{\pi_\emph{y}}$ where $\pi_i$ denotes probability of selecting a $i^{th}$ class instance. Thoretically, $\frac{n_{\emph{y}'}}{n_\emph{y}}=  \frac{\pi_{\emph{y}'}}{\pi_\emph{y}}$. This method provides more degree of freedom compared to fixed power coefficient $\frac{1}{4}$. It provides the capability to fine-tune the $\tau$ parameter based on the interested dataset and the class imbalance ratio.
\begin{equation}
\emph{l}(\emph{y},\emph{f}(\emph{x}))=\text{log}[1+\sum_{\emph{y}'\neq \emph{y}} (\frac{\pi_{\emph{y}'}}{\pi_\emph{y}})^{\tau}.e^{f_{\emph{y}'}(\emph{x})-f_{\emph{y}}(\emph{x})}]
\end{equation}

Since these methods only control the logit distribution, it can result in a diffusion of embeddings. Therefore the learnt embedding distributions must be controlled explicitly. Furthermore. allowing larger margins for the rare classes may result in narrow margins for the frequent classes, therefore sacrificing the class accuracy.
\subsubsection{Loss modification by Multiplicative Logit Adjustment:}

Multiplicative logit adjustment addresses the long-tail classification problem at an angle different to the class margins. Instead, it exploits the issue of overfitting to the minority classes. To explain the accuracy drop in minority classes in terms of the overfitting Ye et al.~\cite{ye2020identifying} show that the learned features $\psi(x)$  corresponds to tail classes during the training and testing are significantly deviated. Therefore, the learned classifier ultimately fails to correctly identify the test instances of the tail classes. In other words, the authors show that this would result in the classifier yielding lower decision values for the test instances. Multiplicative logit adjustment does not explicitly control the learnt feature distribution hence, the training to test feature deviation can not be completely mitigated yet attempts to alleviate the effects of feature deviation by encouraging the tail class decision values to grow larger during the training phase. To elaborate further the idea of the multiplicative logit adjustment is to reduce the drop in decision values from training to test instances.   
To illustrate multiplicative logit adjustment we can discuss the Class-dependent temperatures introduced in~\cite{ye2020identifying} for training a ConvNet. The intuition behind this concept is to mimic the deviation of the features (learned embedding) of the ConvNet.  To this end, the decision values of the minority classes are purposefully reduced to support ConvNet to teach the minority classes more effectively. Temperature for class c is denoted by $\Delta_c$. Where $\Delta_c= {\frac{N_{max}}{N_c}}^\gamma$, $N_c$ is the number of training instances corresponding to class c, $N_{max}$ is the number of training instances corresponding to the vast majority class and $\gamma\geq0$ is a hyperparameter. Hence $\Delta_c=1$ for the vast majority class. The remaining classes are assigned with gradually increasing temperature values beyond one if $\gamma>0$. When $ \gamma =0 $, the objective function reduces to the conventional empirical risk minimization. The multiplicative logit adjustment would be $f_{\emph{y}}(x) \to  \frac{f_{\emph{y}}(x)}{\Delta_y}$
\begin{equation}
\emph{l}(\emph{y},\emph{f}(\emph{x}))=\text{log}[1+\sum_{\emph{y}'\neq \emph{y}} e^{\frac{f_{\emph{y}'}(\emph{x})}{\Delta_{\emph{y}'}}-\frac{f_{\emph{y}}(\emph{x})}{\Delta_{\emph{y}}}} ]
\end{equation}

This method is also not without limitations. The idea of reducing the actual score value through scalar multiplication would only be true with the positive score values. However, in rare cases in the initial training phase, some deep architectures may yield negative scores where the above argument does not hold.
\subsubsection{Loss Modification by Joint Additive and Multiplicative Logit Adjustment:}
\label{joint}
Later, there was research to analyse how much additive and multiplicative adjustment can benefit minority classes when combined. Kini et al.~\cite{kini2021label}  demonstrate that depending on the stage of the training, additive and multiplicative adjustments have different advantages. The authors note that additive adjustments can be beneficial in the initial phase of the training as they can counter the negative effects of the multiplicative adjustment in the initial phases. On the other hand, multiplicative adjustments are necessary to push the decision boundaries towards the majority classes during the terminal phase of training. In this context, the terminal phase of training (TPT) indicates training beyond when the loss has already reached zero. TPT is conducted to obtain sufficient generalization and adversarial robustness for deepnets. Based on these observations, the authors propose a novel vector-scaling (VS) loss that merges additive and multiplicative logit adjustments. The authors claim that margin adjustment during the terminal phase of training can be made effective through multiplicative logit adjustments while additive logit adjustments benefit the initial training phase. Logit modification for the VS loss is denoted by,
$f_{\emph{y}}(x) \to  \frac{f_{\emph{y}}(x)}{{\color{blue}\underbrace{{\Delta_y}}_\text{Multiplicative Adjustment}}} +{\color{purple}\underbrace{\tau ln(\pi_y)}_\text{Additive Adjustment}}$
Equation~\ref{joint} denotes the modified cross entropy loss with the jointly adjusted logit with additive and multiplicative components. The effect of the multiplicative adjustment is experimentally proven to dominate in the terminal phase of training leading to fast convergence. 
\begin{equation}
\emph{l}(\emph{y},\emph{f}(\emph{x}))=\text{log}[1+\sum_{\emph{y}'\neq \emph{y}} e^{\tau ln(\pi_{\emph{y}'})-\tau ln(\pi_\emph{y})} . e^{\frac{f_{\emph{y}'}(\emph{x})}{\Delta_{\emph{y}'}}-\frac{f_{\emph{y}}(\emph{x})}{\Delta_{\emph{y}}}} ]
\label{joint}
\end{equation}

\begin{table}[t]
\centering
\begin{tabular}{lcc}
\toprule
\textbf{Logit Adjustment Technique} & \multicolumn{2}{c}{\textbf{Balanced Accuracy}}\\  \hline
\emph{\textbf{Additive Methods}} & CIFAR 10 &CIFAR100 \\ \hline
 LDAM \cite{cao2019learning} &73.4&39.6\\
 EQL\cite{tan2020equalization}  & 73.8& 39.3\\ 
 LA \cite{menon2020long} &81.2& 43.4  \\ 
 BSoftMax  \cite{ren2020balanced}& 83.1&50.3\\ \hline
\emph{\textbf{Multiplicative Methods}}&&\\ \hline
 CDT~\cite{ye2020identifying}& 84.5& 51.6 \\ \hline
\emph{\textbf{Combined Methods}}&&\\\hline
 VS Loss~\cite{kini2021label}  & 80.82&43.52\\\bottomrule

\end{tabular}
\caption{Summary of the Top-1 accuracy for logit adjustment based  classification for CIFAR 10 and 100 datasets under class imbalance ratio of 100. ResNet-32 architecture is used for both datasets. The accuracy values are taken from the original work cited in the table.}
\label{tab1}
\vspace{-10mm}
\end{table}
\subsubsection{Loss Modification by Angular Margin Adjustments:}
 Instead of using Euclidean distance to measure the difference between features in softmax loss formulation, angles are used as the distance metric. In this scenario, angular margins are exploited. Liu et al.~\cite{liu2017sphereface} propose angular Softmax loss instead of the standard Softmax loss to exploit application-specific inter-class and intra-class distance. This loss learns more discriminative face features compared to standard Softmax loss. The following works described in this section convey different approaches to compute the angular margin. In this context, some works explored the possibility of enforcing margins based on the class rarity under non-stationary data distributions and also applying margins that can transfer the knowledge learnt from the head classes to the tail classes.

The progressive margin loss~\cite{deng2021pml} approach is developed to adjust semantic margin so that the intra-class variance is decreased and the inter-class variance is increased. This method is highly useful when the class order is important (i.e., facial age classification). Most of the existing long-tail methods discard the neighbouring information of the classes. The key components of this approach are 1) ordinal margin learning which extracts discriminative features while preserving the relations of the class order and  2) variational margin learning in which the margin adjusts the decision boundaries of the minority classes through the knowledge transferred from the majority classes. These margins are optimized using the standard backpropagation. Let the output logits of the backbone feature extractor, ordinal margin learning branch and the variational margin learning branch be $f_e(.)$, $f_o(.)$ and $f_v(.)$, respectively. Subsequently, class centre, inter-class variance, and intra-class variance are denoted by $c,\;\mu_1,\;\mu_2$. The loss for a sample (x,y) is given by,
\begin{equation}
l(y,x,W)=\frac{e^{s(x_y,w_y)-m_{py}}}{ e^{s(x_y,w_y)-m_{py}} + \sum_{y’\neq y} e^{s(x_y,w_{y’})}}
\end{equation}
Where $s(.)$ is the similarity function and $m$ represents the learned margin values corresponding to class $y$.$\mathbb{M}_o$ denoted in Equation~\ref{mo} can be optimized combined with the cross entropy minimization and $\mathbb{M}_o^* \in R^{c\text{x}c}$. $\mathbb{M}_o$ improves the feature discriminativeness based on the ordinal relation. $\mathbb{M}_v$ denoted in Equation~\ref{mv} and $\mathbb{M}_o^*$ act complementary to each other by $\mathbb{M}_v$ improves the learned features for each class while decreasing the chance for confusion between classes. Therefore the progressive margin can be defined by $\mathbb{M}_{py} = \lambda\mathbb{M}_o^*+\beta \mathbb{M}_v$

\noindent\begin{minipage}{.5\linewidth}
\begin{equation}
\mathbb{M}_o =f^o([c,\mu_1,\mu_2])
\label{mo}
\end{equation}
\end{minipage}%
\begin{minipage}{.5\linewidth}
\begin{subequations}
\begin{equation}
\Delta V=[c^t,\mu_1^t,\mu_2^t]- [c^{t-1},\mu_1^{t-1},\mu_2^{t-1}]
\end{equation}
\begin{equation}
\mathbb{M}_v= f^v(\Delta V) \text{  } \mathcal{M}_v \in R^c
\label{mv}
\end{equation}
\end{subequations}
\end{minipage}

Wang et al.~\cite{wang2018additive} propose a more comprehensive means to utilize the angular margin in the softmax loss as indicated in Equation~\ref{ang}. Where, $ cos(\theta_{y_i})=\frac{W_y^T \Phi(x)}{ ||W_y^T|||| \Phi(x)||} $. The s is used to scale the cosine values as in the above work which is learnt by backpropagation and m > 1 is an integer which controls the level that the classification boundary could be pushed.
\begin{equation}
L(y,x,W)=-\frac{1}{n} \sum_{i=1}^n \log \frac{e^{s.(\cos\theta_{y_i} -m)}}{ e^{s.(\cos\theta_{y_i} -m)} +\sum_{j=1,j\neq y_i}^c e^{s.\cos\theta_j}}
\label{ang}
\end{equation}

\subsection{Summary}
To summerise, Table~\ref{tab1} compares Top-1 balanced accuracies for different logit adjustment techniques. Angular margin-based methods are not compared as those techniques are mainly implemented for face recognition. According to the table,  balanced softmax and CDT perform superior even when the number of classes grows from 10 to 100. VS loss is the combined additive and multiplicative adjustment loss which underperforms compared to additive loss BSoftMax and multiplicative loss CDT for the CIFAR datasets. Logit adjustment emphasises enforcing a larger margin for the minority classes. It does not take into account optimizing the learnt representation explicitly. Therefore it is important to optimize the learnt representation to increment the performance so that the approach is complementary to the logit adjustment. We discuss the optimized representation learning in the next stage, the ways to use those techniques, and the logit adjustment approaches.

\section{Optimized Representation Learning}
\label{opt}

As it is empirically observed, embeddings for tail classes tend to diffuse among majority class embeddings as the neural models can not explicitly control the distribution of the learned embedding itself. In this section, we discuss different approaches to explicitly optimize the learnt feature distribution as indicated in figure~\ref{tax} Optimized representation learning branch.  Metric learning, contrastive learning,two-staged methods, ensemble learning, and efficient training strategies are discussed in these approaches.

\subsection{Metric Learning Approaches}
\label{ss1}
In~\cite{jitkrittum2022elm}, the authors propose an efficient means to incorporate both logit and embedding margins considering the ideas of metric learning and representation learning. They also demonstrate examples of how the embeddings learned by minimizing the cross entropy with a higher train and test accuracy can still be diffused. The proposed minimization objective to rectify this issue is denoted by,
\begin{equation}
\min_W \frac{1}{N} \sum_{(x,y) \in S} [l(y,f(x,W)) + \lambda \Omega_{pull}(x,y)]
\end{equation}

The idea behind this objective function is to generate more compact embedding for each class. Here, the trade-off between the logit margin and the embedding margin is controlled by parameter $\lambda > 0$. Using the concepts from metric learning, the authors have defined the $\Tilde{\Omega}_{pull}(x,y)$ as follows.

\begin{equation}
\Tilde{\Omega}_{pull}(x,y)= \max_{x^+ \in S_y \backslash \{x\}} [||\Phi(x)-\Phi(x^+))||_2^2-\alpha_y]_+
\end{equation}

 where $\alpha_y$ is the margin for the maximum distance between the embeddings of the example pairs for class $y$, $\alpha_y \propto \pi_y^a$. Here $a>0$ and $\pi_y$ is the probability of class y in the training set. $\Omega_{pull}$ is the differentiable relaxation of the $\Tilde{\Omega}_{pull}(x,y)$
\begin{equation}
\Omega_{pull}(x,y)= \log [1+\sum_{x^+ \in S_y \backslash \{x\}} e^{||\Phi(x)-\Phi(x^+)||_2^2-\alpha_y}]
\label{ob1}
\end{equation}

To overcome the higher level of uncertainty associated with the rare classes, the proposed objective function in Equation~\ref{ob1} forces the model to place all the tail class embeddings more compactly compared to frequent classes due to the corresponding selection of $\alpha_y$. This will implicitly guarantee that the rare class embedding is well separated from the other classes, and it will enhance the classification accuracy for unseen test instances. Therefore, it is crucial to enforce the class-specific embedding margin $\alpha_y$  to ensure sufficient regularization on the tail class embeddings. 

Wang et al.~\cite{wang2019dynamic} also propose another framework to combine cross entropy loss and the metric learning loss during deep CNN training. However, they argue that giving equal weight to the two loss components may not fully utilize the discriminative power of the deep CNN. The authors clarify that the cross entropy loss focuses more on the classification problem through the assignment of correct labels to the input instances, while metric learning works more on learning the soft feature embeddings to represent each class in the feature space distinctively. In the latter case, no label assignment is considered. Here, the authors learn a schedule to manage the training weighting on the two loss components. Through this loss, higher importance is assigned to the metric learning loss during the initial learning stage to provide sufficient freedom to adjust the feature embeddings. In the latter stages, cross entropy loss is given high importance in fine-tuning the classifier. The corresponding semantic description of the additional metric learning loss, which is termed triplet loss with easy anchors, is denoted as follows,

\begin{equation}
L_{TEA}= \frac{\sum_T \text{max}(0,m_j+d(x_{\text{easy,j}},x_{+,j})-d(x_{\text{easy},j},x_{-,j}))}{|T|}
\end{equation}

Here $x_{\text{easy},j}$ denotes an easy positive (with the highest score value) of class j, $x_{+,j}$ is a hard positive of class j and similarly, $x_{-,j}$ is a hard negative of class j. T denotes the number of triplet pairs. $m_j$ is the triplet loss margin for the class $j$. Further, d(.) denotes the pairwise distance of two instances in the feature space. i.e.,
$d(x_{1,j},x_{2,j})= || \Phi(x_{1,j}-\Phi(x_{2,j})||_2$.
Let $g(l)$ be a function that monotonically decreases from 1 to 0, where $l$ denotes the current training epoch. Then,
\begin{equation}
L_{\text{total}} = L_{\text{CE}}+ g(l) * L_{\text{TEA}}
\label{t1}
\end{equation}
\subsection{Contrastive Learning Approaches}
\label{ss2}
More recently, contrastive learning has become popular in learning advanced representations that can be effectively used for various downstream tasks. Supervised contrastive learning (SCL) approaches have been successfully used in large-scale classification problems. Typically, SCL loss attempts to reduce the distance between features generated from the augmentation of the same image and increase the distance between features generated from augmented images from negative pairs. While the SCL technique substantially improves when applied to balanced data, its performance under imbalanced data remains largely experimental. However, experimental evidence has shown that including augmented views associated with SCL significantly enhances accuracy, independent of the data distribution.
In~\cite{zhu2022balanced}, authors propose balanced contrastive learning(BCL) loss to optimize the supervised contrastive loss.  In the setting of long-tail training distribution, where batch-wise class averaging and class complements concepts are used to formulate the new BCL loss denoted in Equation~\ref{bcle}. This loss alleviates the head class bias. Class complement
 ensures that all the classes are considered in every mini-batch by exploiting the class centre prototypes extracted during the learning process. Batch-wise averaging the exponential of class similarity terms will also reduce the bias towards the head classes. A cross entropy loss branch is also implemented parallel to the BCL branch. The combined CE and BCL loss is considered as the total objective loss.
\begin{equation}
L_{\text{BCL}}=-\frac{1}{|B_y|} \cdot \\ \sum\limits_{p\in {B_y \{i\}} \cup \{c_y\}}\log \frac{e^{z_i.z_p}}{\sum\limits_{j\in Y}\frac{1}{|B_j|+1}\sum\limits_{k \in B_j \cup \{c_j\}} e^{z_i.z_k}}
\label{bcle}
\end{equation}

\subsection{Two Staged Training}

Earlier in this section we discussed the importance of classifier learning and representation learning. The works presented in Section~\ref{ss1} and Section~\ref{ss2} demonstrate the case of simultaneous classifier and representation learning through a combined objective. In contrast, Kang et al.~\cite{kang2019decoupling} experimentally validate that decoupling the representation learning and the classifier learning can enhance the overall accuracy for the. In this work, classifier and representation learning are conducted sequentially. In the initial stage, representation is learnt with standard instance-based sampling (sampled based on the original training data distribution) or class balanced sampling (giving equal probability for each class when sampling) or a mixture of them. In stage two, classifier boundaries are balanced on top of the learnt representation with either retraining with class-balanced sampling, post-hoc adjustments or non-parametric methods i.e., nearest class mean classifiers.

Similarly, Cao et al.~\cite{cao2019learning} claim that commonly used re-sampling and reweighting strategies are inferior to the vanilla empirical risk minimization algorithm before applying the annealing rate. Here, the annealing rate is the parameter to control the progressive learning rate decay schedule. The authors propose a deferred re-balancing strategy, which initially trains using the vanilla ERM with the LDAM loss~\cite{cao2019learning} before annealing the learning rate and later implements a reweighted LDAM loss along with a decreased learning rate. This method technically can be coupled with any long-tailed losses described in Section~\ref{logit3}. This strategy is experimentally shown to provide better initialization at the start of the training process to benefit the second stage of training, where the learning rates are lower.

In ~\cite{alshammari2022long}, the authors introduce two two-stage learning frameworks based on simple regularizers to enhance the performance of  recognition. This method is attractive since it does not utilize new losses, deferred network architectures or aggressive data augmentation strategies. Instead, stage one trains the standard CE-based backbone with weight decay. In stage two, classifier boundaries are optimized with a class-balanced loss, weight decay, and MaxNorm regularization. Unlike the L2-norm regularizer which forces the norm value for per-filter weights to value one, MaxNorm relaxes the constraints to allow the movement with the norm ball with a radius of 1. Further Weight decay regularization supports learning smaller weights by pulling the values to the origin. MaxNorm encourages weights to increase within a norm sphere but limits them when the norms grow larger than the radius. Since MaxNorm does not pull the weight towards the origin as weight decay does but instead caps the weight norms within the norm ball, it has improved numerical stability.  Both regularizers tune weight norms dynamically during the training stage. Given that weight decay and MaxNorm are quite diverse techniques,  Alshammari et al.~\cite{alshammari2022long} provide theoretical evidence to show that solving the weight decay problem is a single step of resolving the MAxNorm problem. They also experimentally showcase that the integrated MaxNorm and Weight decay regularization can provide better performance to the trained model. The combined MaxNorm and weight decay solution is indicated by Equation~\ref{maxn}.

\begin{equation}
W^* = \arg \min_W \max_{\gamma \geq 0} \frac{1}{N}\sum_{x \in S} l(y,f(x,W)) + \sum_k \gamma(|| w_k||_2^2-\delta))
\label{maxn}
\end{equation}

Where $\gamma$ is the Karush Kuhn Tucker multiplier and $\delta$ is the L2 norm ball radius. Nonetheless, the two-staged learning can involve tedious hyper-parameter tuning. To this end, Zhang et al.~\cite{zhang2021distribution} propose a novel two-stage approach that can significantly reduce the need for heuristic design requirements in the process. The framework also consists of two stages. In the initial stage, feature extractor $\psi(x)$ and the classifier head $f(x)$ are jointly learned on the original long-tailed data distribution. The second stage is calibration, which does not require additional training. Here, parameters of $\psi(x)$ are frozen, and merely the calibration of the classifier is conducted. For this purpose, an adaptive calibration function and distribution alignment strategy are proposed. The final score for the $i^{th}$ class is given by $f_i(x)$. The authors introduce an adjustment to this score as follows.

\begin{equation}
\bar{f}_i(x)=\alpha_i f_i(x)+ \beta_j  \text{    }\forall i \in L
\end{equation} 

Calibration parameters $\alpha_i$ and $\beta_i$ are learned from data. Authors define  $\sigma(x)$ as a score function that combines the original and adjusted scores.

\begin{equation}
f'_i(x)=\sigma(x).\bar{f}_i(x)+(1-\sigma(x)).f_i(x)\\\\
                     = (1+\sigma(x)\alpha_i).f_i(x)+\sigma(x).\beta_j
\end{equation}

Now, the model prediction obtained from the calibrated score is given by  $p(y=i|x)=\frac{e^{{f'}_i(x)}}{\sum_{k=1}^L e^{{f'}_k(x))}}$. It is denoted as a reference distribution favouring a class-balanced prediction by $p_r(y|x)$. Then, the KL-divergence between the $p(y|x) $ and $p_r(y|x)$ is minimized to align the two distributions using calibration parameters.

\begin{equation}
L_{KL}= \mathbb{E}_S [KL(P_r(y|x)||p(y|x)]\approx \frac{1}{N}\sum_{i=1}^N [\sum_{y=1}^L] P_r(y|x)\log( P(y|x))] + C
\end{equation}

Here, $C$ is a constant. Reference distribution is represented as a weighted empirical distribution of the input data, as indicated in $P_r(y=c|x)= w_c \delta_c(y) \text{  } c \in L$. Where $w_c$ is the class weight and the $\delta_c(y)$ is the Kronecker delta function, equal to 1 when $y=c$ otherwise becomes 0. Let $ \rho $ be a hype-parameter, then $w_c =\frac{\frac{1}{n_c}}{\sum_{k=1}^L \frac{1}{n_k}}$.
According to this interpretation, the reference distribution is known and the model prediction is available after the stage 1 training. Subsequently
  $\alpha_j$ and $\beta_j$  calibration parameters corresponding to each class are learnt from data through the KL divergence minimization. Then calibrated classifier has an adjusted decision boundary for the long-tail case and the corresponding dataset.
The curriculum learning schedule proposed in \cite{wang2019dynamic} is generic in nature. It can be implemented with a combination of CE and metric learning loss and CE and SCL loss. The optimal hyperparameter values are to be found using a grid search.
 Recent work proposed by Du et al.~\cite{cvpr1} employs a straightforward approach to enhance the accuracy of the worst-performing classes. They adopt a two-stage training strategy, wherein the first stage involves training with any existing standard loss function. In the second stage, they freeze the backbone of the model and reinitialize the fully connected layer. Subsequently, they conduct retraining using the proposed Geometric Mean Loss. Harmonic and geometric means are theoretically sensitive to small values; in contrast, arithmetic means are less affected by them.  As harmonic means involves reciprocal, it is numerically unstable and challenging to optimize.  Therefore, in this paper, the geometric mean of per-class recall has been maximized. Refer to the following Equation for the modified loss function.
\begin{equation}
-\frac{1}{L} \sum_{y=1}^{L} \log [{\frac{1}{N_y}}\sum_{j=1}^{N_y}[\frac{N_y e^{f_y(x_j)}}{\sum_{y'=1}^L N_{y'} e^{f_{y'}(x_j)}}]]
\end{equation}
\subsection{Ensemble Learning Approaches}
The main drawback of the state-of-the-art long-tailed methods is they tend to increase the overall and medium shot accuracy with the sacrifice of the many shot accuracy. In~\cite{wang2020long}, authors introduce an ensemble method termed Routing Diverse Experts (RIDE)  to combine different long-tailed techniques so that the many shot medium shot and few shot accuracy s could be increased simultaneously. The framework is also capable of reducing the model variance for all the classes while significantly reducing the model bias for the minority classes. The accuracy of the prediction considering the true value reflects the model bias, the stability of the prediction measures the variance and the irreducible error is independent of the learned model.  This model has a shared set of earlier layers $\psi(x,\theta)$ and the latter independent layers  $f^{\theta_1}(x),f^{\theta_2}(x)…$ set to train based on the selected  techniques. All the layers are jointly optimized.

\begin{equation}
L_{total}= \sum_{i=1}^n (l(y,f^{\theta_i}(\psi(x,\theta))) + \lambda. L_{D-Diversify}(x,y,\theta_i) )
\end{equation}

Regularization $L_{D-Diversify}(x,y,\theta_i) $ for the joint optimization of the experts support the complementary outcomes from the different experts. The regularization maximizes the KL divergence between the expert $i$ and the rest of the experts. However, the main drawback of this method is the joint training of the experts. Yet the KL-divergence support to improve the diversity between experts, it may not be sufficient to train completely independent experts. To alleviate this, Zhang \emph{et al}~\cite{zhang2021self} proposes to train several experts with varied skills that are capable of handling various class distributions. Authors aggregate the experts during the test time so that they can robustly perform well for any unknown test class distribution. They also discuss and experimentally validate the correlation between the learned class distribution and the model loss during the training time.
That is, in a scenario where the training data is long-tailed, the model learned with the softmax loss excels at long-tailed test data classification where the same majority class is the most critical class. ex. Training data consists of vast amounts of COVID-19 diagnosis compared to non-covid patients and test data also includes a large number of COVID positive patients compared to COVID-negative people.
When the state-of-the-art long-tailed methods for the training, the learnt model would perform well for uniform test class distributions. The authors also empirically demonstrate that there is a correlation between model expertise and prediction stability. In other words, stronger experts demonstrate a better prediction similarity between perturbed views of the same instance from the class they excel from. Based on this intuition authors have proposed a self-supervised strategy to maximize the prediction stability through the efficient learning of the aggregation weights of the experts with the unlabeled test data. The prediction stability for an expert is obtained by the cosine similarity of the two predictions obtained by the perturbed vies of an instance of the oncoming test data stream. All the trained experts have a shared backbone $\psi(\theta)$, and subsequently, experts have independent networks E1, E2, and E3. The last layer logit values corresponding to each expert is denoted by $f^1(.),f^2(.)$, and $f^3(.)$. E1 represents the original long-tailed class distribution (based on the softmax loss). The loss component for a training instance is given by,

\begin{equation}
 l_{E1}(y,f(x)) = - \text{log} \frac{e^{f^1_y(x)}}{\sum_{{y’} \in Y} e^{f^1_{y’}(x)}}
\end{equation}

The loss component for E2 is designed to replicate the uniform class distribution by additive logit adjustment as in Equation~\ref{e2}. Inversely, long-tailed class distribution is simulated by learning E3 accordingly. Let $\bar{\pi}$ be the inverted label frequency $\pi$.

\begin{minipage}{0.5\textwidth}
 \begin{equation}
 l_{E2}(y,f(x)) = - \text{log} \frac{e^{f^2_y(x)+\text{log}(\pi_y)}}{\sum_{y’ \in Y} e^{f^2_{y’}(x)+\text{log}(\pi_{y'})}}
 \label{e2}
\end{equation}
\end{minipage}
\begin{minipage}{0.5\textwidth}
  \begin{equation}
   l_{E3}(y,f(x)) = - \text{log} \frac{e^{f^2_y(x)+\text{log}(\pi_y)-\lambda\text{log}(\bar{\pi}_y)}}{\sum_{y’ \in Y} e^{f^2_{y’}(x)+\text{log}(\pi_{y'})-\lambda\text{log}(\bar{\pi}_{y'})}}
\end{equation}
\end{minipage}
The aggregated final prediction is $\bar{y}=\text{SOFTMAX}(a_1.f^1+a_2.f^2+a_3.f^3)$. Where $a_1,a_2$, and $a_3$ are the aggregation weights. The prediction stability is established by maximising the following cosine similarity between the two perturbed inputs of a test instance.
\begin{equation}
\max_{a_1,a_2,a_3} \frac{1}{N_\text{test}} \sum_{S_\text{test}} \bar{y}^1. \bar{y}^2
\end{equation}
\subsection{Efficient Training Strategies}

  Zhong et al.~\cite{zhong2019unequal} propose an iterative way to slowly train the tail data with hard sample mining. The idea is to train the model initially with the head class data with a noise-resistant loss. Secondly, the model will be trained with tail class data to improve inter-class separability information learning. The hard samples are identified and gradually fed to the learning in an iterative manner. This model is claimed to be robust for noisy long-tail learning. VS loss~\cite{kini2021label}, presented in Section~\ref{joint} discusses the terminal phase of training indicating training beyond the zero training error for better convergence. In this scenario, optimization is conducted to push the training loss to zero beyond the zero misclassification error. In this work, authors experimentally and theoretically showcase that even though additive adjustments benefit during the initial phase of training, they are not very effective during the TPT  In contrast, multiplicative logit adjustment performs well during the TPT, enhancing the convergence speed and preserving the accuracy of the minority class. Therefore, the combined VS loss is defined to maximize the use of different loss components during different optimization phases. 
  
In a different approach, Rangwani et al.~\cite{rangwani2022escaping} introduce class-wise hessian analysis to identify the convergence to saddle points. This phenomenon showcases that the accuracy reduction in minority classes can be explained with the saddle points as it results in sub-optimal generalization in minority classes. Authors provide theoretical evidence that sharpness-aware minimization applied to the reweighting method and sufficient regularization successfully escaped saddle points. Sharpness-Aware Minimization target to
 flatten the loss landscape by first identifying a sharp maximal point $\epsilon$ in the selected neighbourhood
of weight parameter w. Subsequently, it minimizes the loss at the sharpest point $(w + \epsilon)$. \\ Ye et al.~\cite{ye2020identifying} experimentally show that there exists a significant scatter of the backbone features when it comes to tail classes. This increases the risk of misclassification for the points close to the feature cluster margin as they may have a high tendency to cross the class boundary. In a recent work by Li et al.~\cite{cvpr2}, Feature cluster compression (FCC) is proposed to increase the backbone feature compactness. The idea is a single-step procedure where backbone features are multiplied by a pre-defined scaling factor before feeding to the classifier as in $\phi_y(x) \to \tau_y\phi_y(x) $. Even though the results does not reach the best SOTA level accuracy the idea is extendable to robust models~\cite{dealvis2023longtail}.
\subsection{Summary}
Table~\ref{fea} denotes the top 1 accuracy of the representation learning techniques discussed in this section. It is clearly visible that modifying the long-tail CE training loss with a BCL loss with augmentation increases the top1 accuracy in $8\%$. SCL and metric learning frameworks optimize the classifier and the feature representation in parallel pipelines. In the two-stage learning methods, representation learning is optimized in the first stage, and the classifier is optimized in the second in a sequential manner. These two-stage methods LWS and LTR also introduce around $8
\% $and $10\%$ accuracy improvement compared to the direct logit adjustment-based method. The significant leap in accuracy is obtained through careful and slow representation learning and sufficient classifier balancing. The multiple expert frameworks also increase the overall performance to a competitive level compared to metric learning, SCL and two-stage learning methods.
\begin{table}[t]
\centering

\begin{tabular}{ccc}
\toprule

\textbf{Method} & CIFAR 100 & CIFAR 10 \\\hline\hline
CE with logit adjustment~\cite{menon2020long} &43.89&77.67\\ \hline
\multicolumn{3}{l}{\textbf{CE with logit adjustment + Metric Learning loss}}\\\hline
 ELM~\cite{jitkrittum2022elm}& 45.77& 77.95\\ \hline
 \multicolumn{3}{l}{\textbf{CE with logit adjustment +  Contrastive Learning loss}}\\\hline
 BCL~\cite{zhu2022balanced} &51.93& 84.32 \\ \hline

\multicolumn{3}{l}{\textbf{Two Stage Learning Methods}}\\ \hline 
 LWS~\cite{kang2019decoupling}& 51.6 &84.5\\ 
 LTR~\cite{alshammari2022long}& 53.5&- \\ \hline
\multicolumn{3}{l}{\textbf{Ensemble Methods}}\\\hline
 SADE~\cite{zhang2021self} & 49.8&-\\
RIDE~\cite{wang2020long}&48.2&-\\\bottomrule
\end{tabular}
\caption{Summary of the top1 accuracy for representation learning techniques based  classification for CIFAR 10 and 100 datasets under class imbalance ratio of 100.ResNet-32 architecture is used and trained for 200 epochs for both datasets }
\label{fea}
\vspace{-10mm}
\end{table}

\section{Balanced Classifier Learning}

The algorithmic level solutions for the classifier balancing mainly come under two categories: Loss Reweighting and Post-hoc adjustments as indicated in Figure~\ref{tax} Balanced classifier learning branch. Loss Reweighting is a popular technique that is used not only in balanced classifier learning but also in sole representation learning and joint classifier and representation learning. This survey discusses all the prominent loss-reweighting techniques in Section~\ref{rew}. Therefore, in this section, we will only focus on the post-hoc adjustment techniques related to balanced classifier learning. 
\subsection{Post-hoc Techniques}
Post-hoc techniques modify the outputs of the model so that it encourages a balanced classifier in the long-tail settings. The most common techniques are post-hoc correction of the decision threshold, normalizing the model weights or adjusting the logits.
\subsubsection{Weight Normalization:} 
\label{nor}
In~\cite{kang2019decoupling} authors state that in a  classification scenario, it is important to adjust the decision boundaries of the jointly learnt classifier (from imbalanced data) during the phase of representation learning. To this end, the authors propose a classifier weight normalisation strategy with a temperature controlled by a hyperparameter. The intuition for this idea is based on the correlation between the norms of the classifier weights and the number of training instances for a certain class when trained with the class imbalanced data. Further, when fine-tuned with class-balanced sampling the classifier weight norms tend to reach the same level of magnitude for all the classes. Based on this phenomenon, directly modifying the classifier weights through the $\tau$ normalisation instead of the class-balanced fine-tuning, to correct the imbalance of the decision boundaries is proposed. Let score function $f_y(x)=w_y\psi(x,\theta)$ where $w_y \in R^d$ are classifier weights of class y. Scaled weights are denoted by $ \tilde{w}_y =\frac{w_y}{||w_y||^\tau}$. $\tau$ is the hyperparameter used to control the normalization temperature. $||. ||$  is the L2 norm. $\tau=1$ denotes the standard L2 normalisation and when $\tau=0$ no scaling is imposed on the weights. After the $\tau$ normalisation the adjusted logit values are given by $f_y(x)\to \frac{f_y(x)}{||w_y||^\tau}$. 

A comparable post hoc adjustment is proposed in~\cite{kim2020adjusting}, where the authors encourage a larger margin for the less frequent classes, allowing sufficient regularization by rescaling the classifier weight norm. The re-scaled weight vector for class $y$ is given  by $ \tilde{w_y} \to {(\frac{N_{max}}{N_y})}^\gamma w_y $. Where $\gamma $ is a tunable hyperparameter, the final logit adjustment can be denoted as follows. $f_y(x) \to {(\frac{n_{max}}{n_y})}^\gamma f_y(x)$

\subsubsection{Logit Modification} 
Post-hoc logit adjustment is necessary for classifiers trained on long-tailed data to adjust the logits based on the nature of the test distribution. This adjustment is usually label-dependent in this context. We have discussed the logit adjustment during training time in Section~\ref{logi}. Similar class frequency-based additive or multiplicative logit adjustment can be adopted in the post-hoc logit adjustment. To make the logit modification more effective, the deepnet may need to be calibrated. The normalization techniques discussed in Section~\ref{nor} can be used to calibrate the model as necessary. Subsequently, the logit adjustments could be made.

In~\cite{menon2020long}, authors train a model to estimate the standard $P(y|x)$ using the standard cross entropy minimization while using long-tailed training data. This model facilitates the post-hoc logit adjustment. Let  $ s^{\ast}: X \to R^L$ be an unknown scorer, it can be assumed that the underlying class probability  $P(y | x) \propto exp({s_y}^{\ast} (x) )$ and by definition $P_{bal}(y | x) \propto P(y | x)/P(y)$ where $s_y$ is the scorer correspond to class y. Therefore, the authors  derive the following equation,
\begin{equation}
\text{argmax}_{y \in [L]} P_{bal}(y | x) = \text{argmax}_{y \in [L]} exp({s_y}^*(x))/P(y) = \text{argmax}_{y\in [L]} {s_y}^*(x)-\text{ln} P(y)
\label{mul}
\end{equation}

According to Equation~\ref{mul}, the score function can be explicitly modified with $lnP(y)$. Another popular approach for post-hoc neural network calibration is Platt scaling~\cite{platt1999probabilistic}. Multi-class extension to Platt scaling applies a linear transformation on the learnt logits~\cite{guo2017calibration}.
Multiplicative and additive terms are applied to the logit values to fine-tune the final prediction.
\begin{equation}
 \bar{y}_i = \text{argmax}_k(Af(x_i)+d)^{(k)}
 \label{plat}
\end{equation}

Plat scaling is not specific to long-tail learning. It adjusts the final probability scores to present a given class's actual likelihood clearly. The intuition for the Plat scaling is to fit a logistic regression model to the predicted probabilities by the trained model. This regression model is learnt on a separate dataset that may be a subset extracted from the training dataset. One of the main drawbacks of this method is its requirement for a separate dataset to learn the calibration parameters. This could be expensive to acquire. $A$ and $d$ can be computed by optimizing negative log-likelihood over the validation set as indicated in Equation~\ref{plat}.  To train a fair classifier (balanced prediction for all the classes), balanced classifier learning is necessary under class imbalance. Yet it may not yield any additional performance improvements by combining margin-based loss modification with the balanced classifier learning techniques as both methods implicitly balance the model prediction~\cite{menon2020long}.

 \subsubsection{Optimal Transport}
 post-hoc corrections with logit adjustment and weight normalizing operate at the individual sample level. Therefore, optimal transport (OT) ideas can be used to ensure the marginal distribution of the entire dataset matches the desired distribution. OT is minimizing the cost of transportation from one probability measure to another. In~\cite{peng2022optimal}, authors enforce this constraint as indicated by $\bar{Y}^T 1_{N}=\mu$. $\bar{Y}$ is the corrected prediction value in the matrix form. $\mu$ is the mean of the desired distribution of the test set. Ideally, corrected prediction distribution must align with the original model prediction distribution. Therefore, the authors use inner-product-based similarity to ensure these requirements. $Y$ is the original model prediction in the matrix form. $C$ denotes some transformation (cost function) to $Y$. Authors select $C$, ensuring it to be positive definite. In this work -log(.) function is selected as the cost function. The standard cost functions could be sub-optimal for real-world datasets. Therefore prediction accuracy can be improved by selecting a better cost function to fit the given long-tail dataset. A significant amount of domain expertise is required to learn the cost function, which is a limitation of this method. 

\begin{equation}
\begin{split}
    \max_{Y}  \langle  C(Y),\bar{Y} \rangle \\
\text{subject to  } 
\bar{Y}^T 1_{N}=\mu \text{  ;  }
\bar{Y} 1_K =1_N
\end{split}
\end{equation}
$\bar{Y}$ is imposed another extra constraint to ensure the standard probability constraints where the sum of the prediction probability over all the classes equals 1. Authors use the Sinkhorn algorithm to compute the adjusted prediction $\bar{Y}$.

\subsection{Summary}

As discussed previously in the survey, representation learning and classifier learning are handled sequentially stage-wise, in parallel pipelines or jointly in a single pipeline. In all three cases, loss reweighting and post hoc corrections can be applied to balance the classifier. post-hoc corrections
can be implemented complementary to reweighting methods~\cite{menon2020long} to obtain an added accuracy improvement. The post-hoc methods applied to the loss reweighting methods introduce and balanced accuracy increment of  $4\%$ approximately for CIFAR10, CIFAR100, ImageNet\_Lt and iNaturalist datasets. The post hoc correction also complements the multi-stage learning cases as well~\cite{kang2019decoupling,alshammari2022long}. However, additive post-hoc correction is redundant to apply along with additive logit adjustment, and the multiplicative logit adjustment may be redundant to apply along with the weight normalizing according to the theoretical discussion presented by Menon et al.~\cite{menon2020long}.

\section{Research Gaps}

The algorithmic level solutions we discuss in this survey under loss reweighting, margin-based logit adjustment, optimized representation learning and balanced classifier learning introduce a significant accuracy increment for long-tail cases. Yet the balanced accuracy for datasets with a higher number of classes such as ImageNet\_Lt tends to remain around $55\%$, even though the class imbalance ratio is not extreme but around 1:256. This phenomenon is evident with CIFAR100\_Lt as well, even with a lower number of classes $(<=100)$. This value is even lower for datasets such as  Places\_Lt, which is close to $40\%$-$50\%$. It remains a challenging issue to increase the many-shot, medium-shot and fewshot accuracy simultaneously. This issue is typically addressed by using multiple ensemble experts for each of the cases which remains sub-optimal. Another key issue is long-tail methods involves with loss modifications may ultimately yield sub-optimal solutions because these modifications may damage the consistency that underpins the soft-max cross entropy. Even the improved reweighting and logit adjustment techniques may also lead to overfitting of the minority classes converging into saddle points. It also has been demonstrated that SGD performs poorly in escaping saddle points in tail classes, unlike the natural class-balanced learning scenarios. Therefore, convergence study is important when implementing class imbalance handling solutions and to ensure the convergence of tail classes. A point of interest is regularization conducted in SOTA methods on the tail classes, which are mainly data-dependent (easy and hard example mining), label-dependent (class frequency-based) or model parameter-dependent (Max-Norm, weight decay). Nevertheless, it is essential to consider the similarity factor between classes when regularising classes is applied during training. i.e., majority classes may need high importance, just as a minority class, due to their less distinctiveness among other classes. It lacks solid research work on this area, especially on determining more and less distinctive classes, whether an ambiguous majority class should be favoured more than a distinct minority class, and how this additional information can be integrated into the other types of regularization techniques.
           
Furthermore, there are more challenging yet practically required versions of the  problem. Firstly, there are different ways to solve the original problem, such as foundation models and fewshot learning. Secondly, handling the class imbalance in an online setting for non-stationary training data distributions is also important. Thirdly, the handling of extremely large class imbalance ratios should be addressed. We will briefly discuss the importance of these three areas in the next sub-sections.
\subsection{Zero Shot Learning In the Context of Long-tail Classification}

With the advances in foundation models~\cite{found}, there is a possibility that long-tail classification can be solved using fewshot and zero-shot~\cite{tail1,tail2,tail3} learning. The fewshot learning capabilities of foundation models have also been demonstrated in NLP tasks using Large Language Models~\cite{tail4,tail5}.  Foundation models are trained on large-scale data, commonly using self-supervision. These models could be adapted to a wide range of downstream applications. SOTA example models consist of BERT~\cite{bert}, GPT-3~\cite{gpt3} and CLIP~\cite{clip} models. Billions of parameters associated with these models provide the capability to adapt to downstream tasks merely with a description of the task (prompt). This behaviour could be useful in long-tail classification as the rarity of the training instances would not affect zero-shot learning or fewshot learning scenarios. However, the total elimination of the long-tail bias could be challenging as the original foundation model may have an inherent bias, depending on the training data. Unfortunately, access to these original training data could be restricted due to copyright concerns. Therefore, compensating for this label bias is not trivial. For classification problems, conducting domain-specific tasks could be challenging for zero-shot models, i.e., identifying the car model and species of trees. As a result, ensemble learning models are proposed to combine zero-shot and fine-tuned (fewshot) models to enhance the classification accuracy~\cite{zf}. It is possible that this approach could be beneficial for the tail class classification with the enforced generalisation.
There are different mechanisms and intuitions behind ZSL. Firstly,  we know the main idea behind zero-shot learning (ZSL) is that a model learns to generalise for unseen classes. ZSL conducts the modelling of the unseen classes from the seen classes through the auxiliary or metainformation available with the data. This could be a description, semantic information, an image caption, or a word embedding. Further, it could also be multimodal sensor data. i.e., in an autonomous vehicle, image classification can be supported by 3D point cloud data and thermal image data. ZSL is a form of supervised learning, yet it does not use actual labels. But rather tries to learn the links, properties and correlations between the training data and the corresponding auxiliary information.
Secondly, Contrastive learning is dynamic in class imbalance handling and zero-shot learning.  ZCL methods attempt to learn a classifier for unseen classes based on their inter-and intra-class relationships. This means the similarities and dissimilarities between a given instance and seen classes can be analysed, and a decision can be made on whether the instance belongs to a seen or unseen class.  Therefore, there is room for improvement to adopt CL ideas integrated with ZSL to the longtail settings.
The third idea is the feasibility of establishing a connection between seen and unseen classes in the semantic space, a high-dimensional vector space where the class prototypes are represented as vectors. In ZSL, this knowledge space from the seen class is transferred to unseen classes. To elaborate further, labelled instances in the feature space and the prototypes represented in the semantic space could be projected into a common space. Subsequently labelled instances for the unseen class could be generated considering the relationships between labelled instances and the prototypes represented in the common space. This approach could be considered as a data augmentation solution to alleviate the underrepresentation of the tail classes. Along this direction, ZSL could support finding data-level solutions to class imbalance problems.

\subsection{Long-tail classification with Unknown or Dynamic Class Imbalance Ratios Under Online Settings }

Typical class imbalance handling methods such as LDAM, BALMS, and additive logit adjustment have impaired performance when a certain batch's exact class imbalance ratio is unknown and dynamic over the oncoming data stream. Existing methods have limited capacity to address that. Nonetheless, there have been some recent works in this area with more opportunities to expand the state-of-the-art further. In~\cite{91} authors proposed Wrapped Cauchy Distributed Angular Softmax.  This approach models the feature representation parametrically. They conduct adaptive label-aware regularization of the class margins to generalize the rare classes sufficiently using a concentration factor instead of the exact class imbalance ratio. \cite{92} also discusses a similar approach where authors propose Heteroskedastic Adaptive Regularization (HAR), which attempts to decide the instance level regularization strength adaptively based on absolute rarity and the noise in the instances by optimizing a Lipschitz-regularized objective. Yet discovering the optimal regularization strength remains challenging due to the practical issues.

When it comes to continuous data streams, some cases demand the necessity to learn from partial(i.e, missing labels for some of the image pixels)/missing labels(s instances, as it is not feasible to obtain labels indefinitely. Under such scenarios, preserving the robustness of class imbalance handling is challenging as the true imbalance is unknown.  According to~\cite{95}, self-supervised (SSL) can perform well with partially labelled data compared to supervised learning (SL) as SSL methods learn richer informative features from the head classes compared to SL. This may help to distinguish the rare classes. Even though SSL can alleviate the class imbalanced effect using the large number of unlabelled training data it still causes a bias towards frequent classes due to the bias in the generated pseudo labels. In \cite{95}, authors propose a Frequency-Aware Prototype Learning stage where prototypes are identified from the unlabelled data, which exhibits the underlying long-tailed data distribution. Then, Prototypical Re-Balanced Self-Supervised Learning is used to train CNN models using unlabeled long-tail data. It is an emerging research area to adapt existing re-balancing and cost-sensitive learning ideas into a setting of online learning.

 As the deep neural network optimization, is a content-aware process, the learnt model gives more bias in memorizing the frequent class and it will tend to forget the rare class more easily. In \cite{96} propose a dynamic self-competitor model to contrast with the target model, where the self-competitor is a pruned version of the target model.  The idea is to assist the target in classifying the tail classes accurately through adaptive online mining of easily forgotten instances. This area yet remains open for further analysis.

Apart from handling class imbalance under non-stationary data distribution, addressing the noisy label problem in parallel to the missing label problem is also mandatory.  In balanced datasets, noisy data can be eliminated from training but this is not feasible with long-tail classification due to the scarcity of labels in tail classes. Meta-learning-based methods ~\cite{400,401} are one of the efforts for handling class-imbalanced learning with noisy labels. However, more accurate noisy label detection methods are required to correctly identify the labels that are prone to noise. In \cite{104}, authors propose a framework that attempts noise detection considering the label distribution. This area is open to more comprehensive research.

\subsection{Classification Under Extreme Class Imbalance} 
One of the main issues with the existing SOTA LTC long-tail datasets is that the maximum usable class imbalance ratio among the classes remains lower. The maximum class imbalance ratio used with CIFAR10, CIFAR100, and ImageNet is in the range of 1:200. The highest class imbalance ratio among the datasets is 1: 26,000, which is with the LVIS data set. It is important that the necessary SOTA datasets be developed with extreme class imbalance scenarios. 
In real-world scenarios such as oncoming passenger screening and fraud detection, the actual class imbalance exceeds
1: $10^7$. Under such imbalances the behaviour of the deepnets for multiclass classification is unknown. It requires experimenting with the compatibility and robustness of existing long-tail handling approaches with deep nets such as Tabnet~\cite{tabnet} used for tabular data with extreme imbalance ratios.

\section{Conclusion}

 In this paper, we surveyed recent advances in deep long-tail classification focusing on algorithmic-level level solutions.  We proposed a taxonomy to categorise the works into four broad categories: 1) Loss reweighting, 2) Margin-based Loss Modification, 3) Optimized Representation Learning, and 4) Balanced Classifier Learning, followed by subsequent sub-categories. We presented these works under a common mathematical notation, making it easier to follow across works and compare their inner working conveniently. In addition, we discuss standard performance metrics such as balanced accuracy and proxy measures such as feature deviation analysis. Finally, we discussed existing research gaps and future research directions in long-tail classification. A notable observation in the literature is SCL loss implemented with data augmentation combined with logit-adjusted CE loss yields significant performance leaps compared to other SOTA methods. Furthermore, we discussed that some long-tail techniques could be redundant to use jointly i.e., using class-balanced sampling for training data along with loss reweighting based on the inverse class frequency and loss-based logit adjustment during training time applied with post hoc logit adjustment.
 Despite advances in machine learning and deep learning, long-tail classification remains a challenge that hasn't yet been resolved completely with performance. This phenomenon is visible regarding SOTA datasets, as an average-level class imbalance would significantly affect the classification accuracy. CIFAR10 shows a balanced accuracy of 99\% in the class balanced case; when IF =100, the best accuracy stays in the range of 85\%-86\% with the best combination of multiple long-tail handling techniques, including both data-level and algorithmic level. Similarly, with CIFAR100, accuracy is 96\%-97\% for the class balanced case, and when IF=100, the best accuracy is in the range of 53\%-54\%. ImageNet has an accuracy of 92\%-93\%, and for ImageNet\_Lt with IF of 256, this value drops to 56\%-57\%. iNaturalist is inherently long-tailed. In this dataset, from lowest IF to highest IF (500), the accuracy drops from 93\%-94\% to 71\%-72\%. These metrics show that although several advanced approaches are practised to handle longtail classification, there is still room for improvement to bring classification accuracy closer to a balanced class case.

\section{Acknowledgments}
This research was conducted by the University of Sydney for National Intelligence Postdoctoral Grant (project number NIPG-2022-006) and funded by the Australian Government.

\bibliographystyle{references/ACM-Reference-Format}
\bibliography{references/sample-base}


\begin{thebibliography}{89}


\ifx \showCODEN    \undefined \def \showCODEN     #1{\unskip}     \fi
\ifx \showDOI      \undefined \def \showDOI       #1{#1}\fi
\ifx \showISBNx    \undefined \def \showISBNx     #1{\unskip}     \fi
\ifx \showISBNxiii \undefined \def \showISBNxiii  #1{\unskip}     \fi
\ifx \showISSN     \undefined \def \showISSN      #1{\unskip}     \fi
\ifx \showLCCN     \undefined \def \showLCCN      #1{\unskip}     \fi
\ifx \shownote     \undefined \def \shownote      #1{#1}          \fi
\ifx \showarticletitle \undefined \def \showarticletitle #1{#1}   \fi
\ifx \showURL      \undefined \def \showURL       {\relax}        \fi
\providecommand\bibfield[2]{#2}
\providecommand\bibinfo[2]{#2}
\providecommand\natexlab[1]{#1}
\providecommand\showeprint[2][]{arXiv:#2}

\bibitem[Alshammari et~al\mbox{.}(2022)]%
        {alshammari2022long}
\bibfield{author}{\bibinfo{person}{Shaden Alshammari}, \bibinfo{person}{Yu-Xiong Wang}, \bibinfo{person}{Deva Ramanan}, {and} \bibinfo{person}{Shu Kong}.} \bibinfo{year}{2022}\natexlab{}.
\newblock \showarticletitle{Long-tailed recognition via weight balancing}. In \bibinfo{booktitle}{\emph{Proceedings of the IEEE/CVF Conference on Computer Vision and Pattern Recognition}}. \bibinfo{pages}{6897--6907}.
\newblock


\bibitem[Alvis et~al\mbox{.}(2023)]%
        {dealvis2023longtail}
\bibfield{author}{\bibinfo{person}{Charika~De Alvis}, \bibinfo{person}{Dishanika Denipitiyage}, {and} \bibinfo{person}{Suranga Seneviratne}.} \bibinfo{year}{2023}\natexlab{}.
\newblock \bibinfo{title}{Long-Tail Learning with Rebalanced Contrastive Loss}.
\newblock
\newblock
\showeprint[arxiv]{2312.01753}~[cs.LG]


\bibitem[Arik and Pfister(2021)]%
        {tabnet}
\bibfield{author}{\bibinfo{person}{Sercan~{\"O} Arik} {and} \bibinfo{person}{Tomas Pfister}.} \bibinfo{year}{2021}\natexlab{}.
\newblock \showarticletitle{Tabnet: Attentive interpretable tabular learning}. In \bibinfo{booktitle}{\emph{Proceedings of the AAAI conference on artificial intelligence}}, Vol.~\bibinfo{volume}{35}. \bibinfo{pages}{6679--6687}.
\newblock


\bibitem[Bartlett et~al\mbox{.}(2017)]%
        {baret}
\bibfield{author}{\bibinfo{person}{Peter~L Bartlett}, \bibinfo{person}{Dylan~J Foster}, {and} \bibinfo{person}{Matus~J Telgarsky}.} \bibinfo{year}{2017}\natexlab{}.
\newblock \showarticletitle{Spectrally-normalized margin bounds for neural networks}.
\newblock \bibinfo{journal}{\emph{Advances in neural information processing systems}}  \bibinfo{volume}{30} (\bibinfo{year}{2017}).
\newblock


\bibitem[Bedi et~al\mbox{.}(2020)]%
        {intrusion}
\bibfield{author}{\bibinfo{person}{Punam Bedi}, \bibinfo{person}{Neha Gupta}, {and} \bibinfo{person}{Vinita Jindal}.} \bibinfo{year}{2020}\natexlab{}.
\newblock \showarticletitle{Siam-IDS: Handling class imbalance problem in intrusion detection systems using siamese neural network}.
\newblock \bibinfo{journal}{\emph{Procedia Computer Science}}  \bibinfo{volume}{171} (\bibinfo{year}{2020}), \bibinfo{pages}{780--789}.
\newblock


\bibitem[Bommasani et~al\mbox{.}(2021)]%
        {found}
\bibfield{author}{\bibinfo{person}{Rishi Bommasani}, \bibinfo{person}{Drew~A Hudson}, \bibinfo{person}{Ehsan Adeli}, \bibinfo{person}{Russ Altman}, \bibinfo{person}{Simran Arora}, \bibinfo{person}{Sydney von Arx}, \bibinfo{person}{Michael~S Bernstein}, \bibinfo{person}{Jeannette Bohg}, \bibinfo{person}{Antoine Bosselut}, \bibinfo{person}{Emma Brunskill}, {et~al\mbox{.}}} \bibinfo{year}{2021}\natexlab{}.
\newblock \showarticletitle{On the opportunities and risks of foundation models}.
\newblock \bibinfo{journal}{\emph{arXiv preprint arXiv:2108.07258}} (\bibinfo{year}{2021}).
\newblock


\bibitem[Brown et~al\mbox{.}(2020)]%
        {gpt3}
\bibfield{author}{\bibinfo{person}{Tom Brown}, \bibinfo{person}{Benjamin Mann}, \bibinfo{person}{Nick Ryder}, \bibinfo{person}{Melanie Subbiah}, \bibinfo{person}{Jared~D Kaplan}, \bibinfo{person}{Prafulla Dhariwal}, \bibinfo{person}{Arvind Neelakantan}, \bibinfo{person}{Pranav Shyam}, \bibinfo{person}{Girish Sastry}, \bibinfo{person}{Amanda Askell}, {et~al\mbox{.}}} \bibinfo{year}{2020}\natexlab{}.
\newblock \showarticletitle{Language models are few-shot learners}.
\newblock \bibinfo{journal}{\emph{Advances in neural information processing systems}}  \bibinfo{volume}{33} (\bibinfo{year}{2020}), \bibinfo{pages}{1877--1901}.
\newblock


\bibitem[Buda et~al\mbox{.}(2018)]%
        {survey14}
\bibfield{author}{\bibinfo{person}{Mateusz Buda}, \bibinfo{person}{Atsuto Maki}, {and} \bibinfo{person}{Maciej~A Mazurowski}.} \bibinfo{year}{2018}\natexlab{}.
\newblock \showarticletitle{A systematic study of the class imbalance problem in convolutional neural networks}.
\newblock \bibinfo{journal}{\emph{Neural networks}}  \bibinfo{volume}{106} (\bibinfo{year}{2018}), \bibinfo{pages}{249--259}.
\newblock


\bibitem[Byrd and Lipton(2019)]%
        {byrd2019effect}
\bibfield{author}{\bibinfo{person}{Jonathon Byrd} {and} \bibinfo{person}{Zachary Lipton}.} \bibinfo{year}{2019}\natexlab{}.
\newblock \showarticletitle{What is the effect of importance weighting in deep learning?}. In \bibinfo{booktitle}{\emph{International conference on machine learning}}. PMLR, \bibinfo{pages}{872--881}.
\newblock


\bibitem[Cao et~al\mbox{.}(2020)]%
        {92}
\bibfield{author}{\bibinfo{person}{Kaidi Cao}, \bibinfo{person}{Yining Chen}, \bibinfo{person}{Junwei Lu}, \bibinfo{person}{Nikos Arechiga}, \bibinfo{person}{Adrien Gaidon}, {and} \bibinfo{person}{Tengyu Ma}.} \bibinfo{year}{2020}\natexlab{}.
\newblock \showarticletitle{Heteroskedastic and imbalanced deep learning with adaptive regularization}.
\newblock \bibinfo{journal}{\emph{arXiv preprint arXiv:2006.15766}} (\bibinfo{year}{2020}).
\newblock


\bibitem[Cao et~al\mbox{.}(2019)]%
        {cao2019learning}
\bibfield{author}{\bibinfo{person}{Kaidi Cao}, \bibinfo{person}{Colin Wei}, \bibinfo{person}{Adrien Gaidon}, \bibinfo{person}{Nikos Arechiga}, {and} \bibinfo{person}{Tengyu Ma}.} \bibinfo{year}{2019}\natexlab{}.
\newblock \showarticletitle{Learning imbalanced datasets with label-distribution-aware margin loss}.
\newblock \bibinfo{journal}{\emph{Advances in neural information processing systems}}  \bibinfo{volume}{32} (\bibinfo{year}{2019}).
\newblock


\bibitem[Chawla et~al\mbox{.}(2002)]%
        {126}
\bibfield{author}{\bibinfo{person}{Nitesh~V Chawla}, \bibinfo{person}{Kevin~W Bowyer}, \bibinfo{person}{Lawrence~O Hall}, {and} \bibinfo{person}{W~Philip Kegelmeyer}.} \bibinfo{year}{2002}\natexlab{}.
\newblock \showarticletitle{SMOTE: synthetic minority over-sampling technique}.
\newblock \bibinfo{journal}{\emph{Journal of artificial intelligence research}} (\bibinfo{year}{2002}).
\newblock


\bibitem[Chhabra et~al\mbox{.}(2024)]%
        {tail5}
\bibfield{author}{\bibinfo{person}{Anshuman Chhabra}, \bibinfo{person}{Hadi Askari}, {and} \bibinfo{person}{Prasant Mohapatra}.} \bibinfo{year}{2024}\natexlab{}.
\newblock \showarticletitle{Revisiting Zero-Shot Abstractive Summarization in the Era of Large Language Models from the Perspective of Position Bias}.
\newblock \bibinfo{journal}{\emph{arXiv preprint arXiv:2401.01989}} (\bibinfo{year}{2024}).
\newblock


\bibitem[Collell et~al\mbox{.}(2016)]%
        {collell2016reviving}
\bibfield{author}{\bibinfo{person}{Guillem Collell}, \bibinfo{person}{Drazen Prelec}, {and} \bibinfo{person}{Kaustubh Patil}.} \bibinfo{year}{2016}\natexlab{}.
\newblock \showarticletitle{Reviving threshold-moving: a simple plug-in bagging ensemble for binary and multiclass imbalanced data}.
\newblock \bibinfo{journal}{\emph{arXiv preprint arXiv:1606.08698}} (\bibinfo{year}{2016}).
\newblock


\bibitem[Cui et~al\mbox{.}(2019)]%
        {cui2019class}
\bibfield{author}{\bibinfo{person}{Yin Cui}, \bibinfo{person}{Menglin Jia}, \bibinfo{person}{Tsung-Yi Lin}, \bibinfo{person}{Yang Song}, {and} \bibinfo{person}{Serge Belongie}.} \bibinfo{year}{2019}\natexlab{}.
\newblock \showarticletitle{Class-balanced loss based on effective number of samples}. In \bibinfo{booktitle}{\emph{Proceedings of the IEEE/CVF conference on computer vision and pattern recognition}}. \bibinfo{pages}{9268--9277}.
\newblock


\bibitem[De~Alvis et~al\mbox{.}(2023)]%
        {de2023long}
\bibfield{author}{\bibinfo{person}{Charika De~Alvis}, \bibinfo{person}{Dishanika Denipitiyage}, {and} \bibinfo{person}{Suranga Seneviratne}.} \bibinfo{year}{2023}\natexlab{}.
\newblock \showarticletitle{Long-Tail Learning with Rebalanced Contrastive Loss}.
\newblock \bibinfo{journal}{\emph{arXiv preprint arXiv:2312.01753}} (\bibinfo{year}{2023}).
\newblock


\bibitem[Deng et~al\mbox{.}(2021)]%
        {deng2021pml}
\bibfield{author}{\bibinfo{person}{Zongyong Deng}, \bibinfo{person}{Hao Liu}, \bibinfo{person}{Yaoxing Wang}, \bibinfo{person}{Chenyang Wang}, \bibinfo{person}{Zekuan Yu}, {and} \bibinfo{person}{Xuehong Sun}.} \bibinfo{year}{2021}\natexlab{}.
\newblock \showarticletitle{Pml: Progressive margin loss for long-tailed age classification}. In \bibinfo{booktitle}{\emph{Proceedings of the IEEE/CVF Conference on Computer Vision and Pattern Recognition}}. \bibinfo{pages}{10503--10512}.
\newblock


\bibitem[Devlin et~al\mbox{.}(2018)]%
        {bert}
\bibfield{author}{\bibinfo{person}{Jacob Devlin}, \bibinfo{person}{Ming-Wei Chang}, \bibinfo{person}{Kenton Lee}, {and} \bibinfo{person}{Kristina Toutanova}.} \bibinfo{year}{2018}\natexlab{}.
\newblock \showarticletitle{Bert: Pre-training of deep bidirectional transformers for language understanding}.
\newblock \bibinfo{journal}{\emph{arXiv preprint arXiv:1810.04805}} (\bibinfo{year}{2018}).
\newblock


\bibitem[Di~Martino et~al\mbox{.}(2012)]%
        {fraud}
\bibfield{author}{\bibinfo{person}{Mat{\'\i}as Di~Martino}, \bibinfo{person}{Federico Decia}, \bibinfo{person}{Juan Molinelli}, {and} \bibinfo{person}{Alicia Fern{\'a}ndez}.} \bibinfo{year}{2012}\natexlab{}.
\newblock \showarticletitle{Improving Electric Fraud Detection using Class Imbalance Strategies.}. In \bibinfo{booktitle}{\emph{ICPRAM (2)}}. \bibinfo{pages}{135--141}.
\newblock


\bibitem[Ding et~al\mbox{.}(2021)]%
        {pedes}
\bibfield{author}{\bibinfo{person}{Mengyuan Ding}, \bibinfo{person}{Shanshan Zhang}, {and} \bibinfo{person}{Jian Yang}.} \bibinfo{year}{2021}\natexlab{}.
\newblock \showarticletitle{Improving pedestrian detection from a long-tailed domain perspective}. In \bibinfo{booktitle}{\emph{Proceedings of the 29th ACM International Conference on Multimedia}}. \bibinfo{pages}{2918--2926}.
\newblock


\bibitem[Du and Wu(2023)]%
        {cvpr1}
\bibfield{author}{\bibinfo{person}{Yingxiao Du} {and} \bibinfo{person}{Jianxin Wu}.} \bibinfo{year}{2023}\natexlab{}.
\newblock \showarticletitle{No One Left Behind: Improving the Worst Categories in Long-Tailed Learning}. In \bibinfo{booktitle}{\emph{Proceedings of the IEEE/CVF Conference on Computer Vision and Pattern Recognition}}. \bibinfo{pages}{15804--15813}.
\newblock


\bibitem[Estabrooks et~al\mbox{.}(2004)]%
        {127}
\bibfield{author}{\bibinfo{person}{Andrew Estabrooks}, \bibinfo{person}{Taeho Jo}, {and} \bibinfo{person}{Nathalie Japkowicz}.} \bibinfo{year}{2004}\natexlab{}.
\newblock \showarticletitle{A multiple resampling method for learning from imbalanced data sets}.
\newblock \bibinfo{journal}{\emph{Computational intelligence}} \bibinfo{volume}{20}, \bibinfo{number}{1} (\bibinfo{year}{2004}), \bibinfo{pages}{18--36}.
\newblock


\bibitem[Fan et~al\mbox{.}(2017)]%
        {fan2017learning}
\bibfield{author}{\bibinfo{person}{Yanbo Fan}, \bibinfo{person}{Siwei Lyu}, \bibinfo{person}{Yiming Ying}, {and} \bibinfo{person}{Baogang Hu}.} \bibinfo{year}{2017}\natexlab{}.
\newblock \showarticletitle{Learning with average top-k loss}.
\newblock \bibinfo{journal}{\emph{Advances in neural information processing systems}}  \bibinfo{volume}{30} (\bibinfo{year}{2017}).
\newblock


\bibitem[Fang et~al\mbox{.}(2023)]%
        {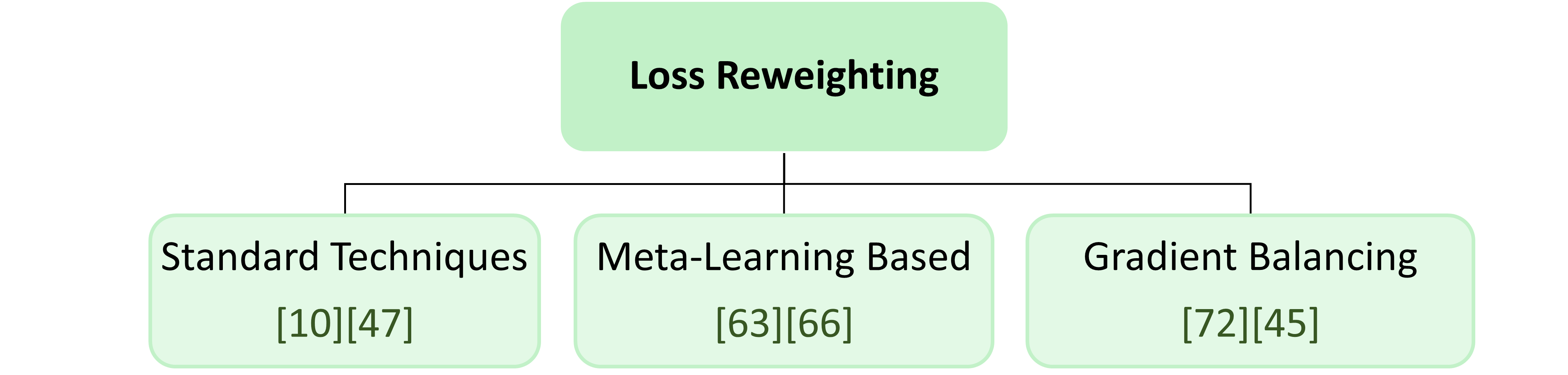}
\bibfield{author}{\bibinfo{person}{Chaowei Fang}, \bibinfo{person}{Dingwen Zhang}, \bibinfo{person}{Wen Zheng}, \bibinfo{person}{Xue Li}, \bibinfo{person}{Le Yang}, \bibinfo{person}{Lechao Cheng}, {and} \bibinfo{person}{Junwei Han}.} \bibinfo{year}{2023}\natexlab{}.
\newblock \showarticletitle{Revisiting Long-tailed Image Classification: Survey and Benchmarks with New Evaluation Metrics}.
\newblock \bibinfo{journal}{\emph{arXiv preprint arXiv:2302.01507}} (\bibinfo{year}{2023}).
\newblock


\bibitem[Fu et~al\mbox{.}(2022)]%
        {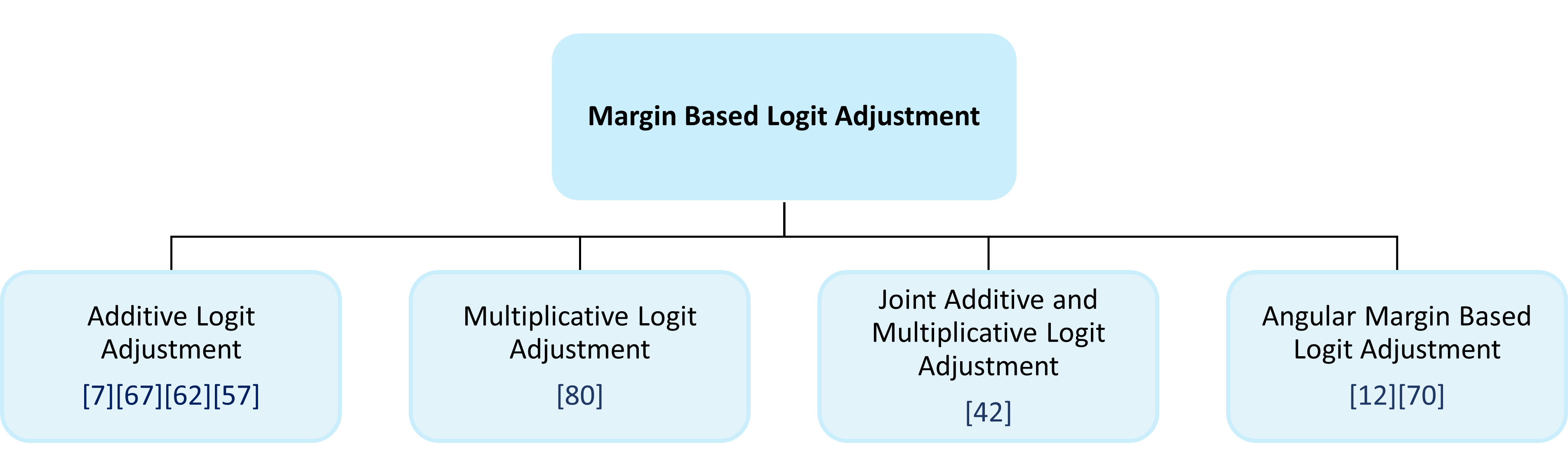}
\bibfield{author}{\bibinfo{person}{Yu Fu}, \bibinfo{person}{Liuyu Xiang}, \bibinfo{person}{Yumna Zahid}, \bibinfo{person}{Guiguang Ding}, \bibinfo{person}{Tao Mei}, \bibinfo{person}{Qiang Shen}, {and} \bibinfo{person}{Jungong Han}.} \bibinfo{year}{2022}\natexlab{}.
\newblock \showarticletitle{Long-tailed visual recognition with deep models: A methodological survey and evaluation}.
\newblock \bibinfo{journal}{\emph{Neurocomputing}} (\bibinfo{year}{2022}).
\newblock


\bibitem[Ghorbani et~al\mbox{.}(2019)]%
        {ghor}
\bibfield{author}{\bibinfo{person}{Behrooz Ghorbani}, \bibinfo{person}{Shankar Krishnan}, {and} \bibinfo{person}{Ying Xiao}.} \bibinfo{year}{2019}\natexlab{}.
\newblock \showarticletitle{An investigation into neural net optimization via hessian eigenvalue density}. In \bibinfo{booktitle}{\emph{International Conference on Machine Learning}}. PMLR, \bibinfo{pages}{2232--2241}.
\newblock


\bibitem[Guo et~al\mbox{.}(2017)]%
        {guo2017calibration}
\bibfield{author}{\bibinfo{person}{Chuan Guo}, \bibinfo{person}{Geoff Pleiss}, \bibinfo{person}{Yu Sun}, {and} \bibinfo{person}{Kilian~Q Weinberger}.} \bibinfo{year}{2017}\natexlab{}.
\newblock \showarticletitle{On calibration of modern neural networks}. In \bibinfo{booktitle}{\emph{International conference on machine learning}}. PMLR, \bibinfo{pages}{1321--1330}.
\newblock


\bibitem[Gupta et~al\mbox{.}(2019)]%
        {gupta2019lvis}
\bibfield{author}{\bibinfo{person}{Agrim Gupta}, \bibinfo{person}{Piotr Dollar}, {and} \bibinfo{person}{Ross Girshick}.} \bibinfo{year}{2019}\natexlab{}.
\newblock \showarticletitle{Lvis: A dataset for large vocabulary instance segmentation}. In \bibinfo{booktitle}{\emph{Proceedings of the IEEE/CVF conference on computer vision and pattern recognition}}. \bibinfo{pages}{5356--5364}.
\newblock


\bibitem[Han(2023)]%
        {91}
\bibfield{author}{\bibinfo{person}{Boran Han}.} \bibinfo{year}{2023}\natexlab{}.
\newblock \showarticletitle{Wrapped Cauchy Distributed Angular Softmax for Long-Tailed Visual Recognition}.
\newblock \bibinfo{journal}{\emph{arXiv preprint arXiv:2305.18732}} (\bibinfo{year}{2023}).
\newblock


\bibitem[He and Garcia(2009)]%
        {survey9}
\bibfield{author}{\bibinfo{person}{Haibo He} {and} \bibinfo{person}{Edwardo~A Garcia}.} \bibinfo{year}{2009}\natexlab{}.
\newblock \showarticletitle{Learning from imbalanced data}.
\newblock \bibinfo{journal}{\emph{IEEE Transactions on knowledge and data engineering}} \bibinfo{volume}{21}, \bibinfo{number}{9} (\bibinfo{year}{2009}), \bibinfo{pages}{1263--1284}.
\newblock


\bibitem[Hong et~al\mbox{.}(2021)]%
        {hong2021disentangling}
\bibfield{author}{\bibinfo{person}{Youngkyu Hong}, \bibinfo{person}{Seungju Han}, \bibinfo{person}{Kwanghee Choi}, \bibinfo{person}{Seokjun Seo}, \bibinfo{person}{Beomsu Kim}, {and} \bibinfo{person}{Buru Chang}.} \bibinfo{year}{2021}\natexlab{}.
\newblock \showarticletitle{Disentangling label distribution for long-tailed visual recognition}. In \bibinfo{booktitle}{\emph{Proceedings of the IEEE/CVF conference on computer vision and pattern recognition}}. \bibinfo{pages}{6626--6636}.
\newblock


\bibitem[Huang et~al\mbox{.}(2019)]%
        {huang2019deep}
\bibfield{author}{\bibinfo{person}{Chen Huang}, \bibinfo{person}{Yining Li}, \bibinfo{person}{Chen~Change Loy}, {and} \bibinfo{person}{Xiaoou Tang}.} \bibinfo{year}{2019}\natexlab{}.
\newblock \showarticletitle{Deep imbalanced learning for face recognition and attribute prediction}.
\newblock \bibinfo{journal}{\emph{IEEE transactions on pattern analysis and machine intelligence}} \bibinfo{volume}{42}, \bibinfo{number}{11} (\bibinfo{year}{2019}), \bibinfo{pages}{2781--2794}.
\newblock


\bibitem[{iNaturalist} 2018 competition dataset(2018)]%
        {iNaturalist18}
{iNaturalist} 2018 competition dataset \bibinfo{year}{2018}\natexlab{}.
\newblock \bibinfo{title}{{iNaturalist} 2018 competition dataset.}
\newblock \bibinfo{howpublished}{~\url{https://github.com/visipedia/inat_comp/tree/master/2018}}.
\newblock


\bibitem[Iranmehr et~al\mbox{.}(2019)]%
        {iranmehr2019cost}
\bibfield{author}{\bibinfo{person}{Arya Iranmehr}, \bibinfo{person}{Hamed Masnadi-Shirazi}, {and} \bibinfo{person}{Nuno Vasconcelos}.} \bibinfo{year}{2019}\natexlab{}.
\newblock \showarticletitle{Cost-sensitive support vector machines}.
\newblock \bibinfo{journal}{\emph{Neurocomputing}}  \bibinfo{volume}{343} (\bibinfo{year}{2019}), \bibinfo{pages}{50--64}.
\newblock


\bibitem[Jiang et~al\mbox{.}(2021)]%
        {96}
\bibfield{author}{\bibinfo{person}{Ziyu Jiang}, \bibinfo{person}{Tianlong Chen}, \bibinfo{person}{Bobak~J Mortazavi}, {and} \bibinfo{person}{Zhangyang Wang}.} \bibinfo{year}{2021}\natexlab{}.
\newblock \showarticletitle{Self-damaging contrastive learning}. In \bibinfo{booktitle}{\emph{International Conference on Machine Learning}}. PMLR, \bibinfo{pages}{4927--4939}.
\newblock


\bibitem[Jitkrittum et~al\mbox{.}(2022)]%
        {jitkrittum2022elm}
\bibfield{author}{\bibinfo{person}{Wittawat Jitkrittum}, \bibinfo{person}{Aditya~Krishna Menon}, \bibinfo{person}{Ankit~Singh Rawat}, {and} \bibinfo{person}{Sanjiv Kumar}.} \bibinfo{year}{2022}\natexlab{}.
\newblock \showarticletitle{ELM: Embedding and Logit Margins for Long-Tail Learning}.
\newblock \bibinfo{journal}{\emph{arXiv preprint arXiv:2204.13208}} (\bibinfo{year}{2022}).
\newblock


\bibitem[Johnson and Khoshgoftaar(2019)]%
        {survey11}
\bibfield{author}{\bibinfo{person}{Justin~M Johnson} {and} \bibinfo{person}{Taghi~M Khoshgoftaar}.} \bibinfo{year}{2019}\natexlab{}.
\newblock \showarticletitle{Survey on deep learning with class imbalance}.
\newblock \bibinfo{journal}{\emph{Journal of Big Data}} \bibinfo{volume}{6}, \bibinfo{number}{1} (\bibinfo{year}{2019}), \bibinfo{pages}{1--54}.
\newblock


\bibitem[Kang et~al\mbox{.}(2020)]%
        {balanced}
\bibfield{author}{\bibinfo{person}{Bingyi Kang}, \bibinfo{person}{Yu Li}, \bibinfo{person}{Sa Xie}, \bibinfo{person}{Zehuan Yuan}, {and} \bibinfo{person}{Jiashi Feng}.} \bibinfo{year}{2020}\natexlab{}.
\newblock \showarticletitle{Exploring balanced feature spaces for representation learning}. In \bibinfo{booktitle}{\emph{International Conference on Learning Representations}}.
\newblock


\bibitem[Kang et~al\mbox{.}(2019)]%
        {kang2019decoupling}
\bibfield{author}{\bibinfo{person}{Bingyi Kang}, \bibinfo{person}{Saining Xie}, \bibinfo{person}{Marcus Rohrbach}, \bibinfo{person}{Zhicheng Yan}, \bibinfo{person}{Albert Gordo}, \bibinfo{person}{Jiashi Feng}, {and} \bibinfo{person}{Yannis Kalantidis}.} \bibinfo{year}{2019}\natexlab{}.
\newblock \showarticletitle{Decoupling representation and classifier for long-tailed recognition}.
\newblock \bibinfo{journal}{\emph{arXiv preprint arXiv:1910.09217}} (\bibinfo{year}{2019}).
\newblock


\bibitem[Kim and Kim(2020)]%
        {kim2020adjusting}
\bibfield{author}{\bibinfo{person}{Byungju Kim} {and} \bibinfo{person}{Junmo Kim}.} \bibinfo{year}{2020}\natexlab{}.
\newblock \showarticletitle{Adjusting decision boundary for class imbalanced learning}.
\newblock \bibinfo{journal}{\emph{IEEE Access}}  \bibinfo{volume}{8} (\bibinfo{year}{2020}), \bibinfo{pages}{81674--81685}.
\newblock


\bibitem[Kini et~al\mbox{.}(2021)]%
        {kini2021label}
\bibfield{author}{\bibinfo{person}{Ganesh~Ramachandra Kini}, \bibinfo{person}{Orestis Paraskevas}, \bibinfo{person}{Samet Oymak}, {and} \bibinfo{person}{Christos Thrampoulidis}.} \bibinfo{year}{2021}\natexlab{}.
\newblock \showarticletitle{Label-imbalanced and group-sensitive classification under overparameterization}.
\newblock \bibinfo{journal}{\emph{Advances in Neural Information Processing Systems}}  \bibinfo{volume}{34} (\bibinfo{year}{2021}), \bibinfo{pages}{18970--18983}.
\newblock


\bibitem[Krawczyk(2016)]%
        {survey8}
\bibfield{author}{\bibinfo{person}{Bartosz Krawczyk}.} \bibinfo{year}{2016}\natexlab{}.
\newblock \showarticletitle{Learning from imbalanced data: open challenges and future directions}.
\newblock \bibinfo{journal}{\emph{Progress in Artificial Intelligence}} \bibinfo{volume}{5}, \bibinfo{number}{4} (\bibinfo{year}{2016}), \bibinfo{pages}{221--232}.
\newblock


\bibitem[Leevy et~al\mbox{.}(2018)]%
        {survey13}
\bibfield{author}{\bibinfo{person}{Joffrey~L Leevy}, \bibinfo{person}{Taghi~M Khoshgoftaar}, \bibinfo{person}{Richard~A Bauder}, {and} \bibinfo{person}{Naeem Seliya}.} \bibinfo{year}{2018}\natexlab{}.
\newblock \showarticletitle{A survey on addressing high-class imbalance in big data}.
\newblock \bibinfo{journal}{\emph{Journal of Big Data}} \bibinfo{volume}{5}, \bibinfo{number}{1} (\bibinfo{year}{2018}), \bibinfo{pages}{1--30}.
\newblock


\bibitem[Li et~al\mbox{.}(2019)]%
        {li2019gradient}
\bibfield{author}{\bibinfo{person}{Buyu Li}, \bibinfo{person}{Yu Liu}, {and} \bibinfo{person}{Xiaogang Wang}.} \bibinfo{year}{2019}\natexlab{}.
\newblock \showarticletitle{Gradient harmonized single-stage detector}. In \bibinfo{booktitle}{\emph{Proceedings of the AAAI conference on artificial intelligence}}.
\newblock


\bibitem[Li et~al\mbox{.}(2023)]%
        {cvpr2}
\bibfield{author}{\bibinfo{person}{Jian Li}, \bibinfo{person}{Ziyao Meng}, \bibinfo{person}{Daqian Shi}, \bibinfo{person}{Rui Song}, \bibinfo{person}{Xiaolei Diao}, \bibinfo{person}{Jingwen Wang}, {and} \bibinfo{person}{Hao Xu}.} \bibinfo{year}{2023}\natexlab{}.
\newblock \showarticletitle{FCC: Feature Clusters Compression for Long-Tailed Visual Recognition}. In \bibinfo{booktitle}{\emph{Proceedings of the IEEE/CVF Conference on Computer Vision and Pattern Recognition}}. \bibinfo{pages}{24080--24089}.
\newblock


\bibitem[Lin et~al\mbox{.}(2017)]%
        {lin2017focal}
\bibfield{author}{\bibinfo{person}{Tsung-Yi Lin}, \bibinfo{person}{Priya Goyal}, \bibinfo{person}{Ross Girshick}, \bibinfo{person}{Kaiming He}, {and} \bibinfo{person}{Piotr Doll{\'a}r}.} \bibinfo{year}{2017}\natexlab{}.
\newblock \showarticletitle{Focal loss for dense object detection}. In \bibinfo{booktitle}{\emph{Proceedings of the IEEE international conference on computer vision}}. \bibinfo{pages}{2980--2988}.
\newblock


\bibitem[Ling and Sheng(2008)]%
        {survey7}
\bibfield{author}{\bibinfo{person}{Charles~X Ling} {and} \bibinfo{person}{Victor~S Sheng}.} \bibinfo{year}{2008}\natexlab{}.
\newblock \showarticletitle{Cost-sensitive learning and the class imbalance problem}.
\newblock \bibinfo{journal}{\emph{Encyclopedia of machine learning}}  \bibinfo{volume}{2011} (\bibinfo{year}{2008}).
\newblock


\bibitem[Liu et~al\mbox{.}(2021)]%
        {95}
\bibfield{author}{\bibinfo{person}{Hong Liu}, \bibinfo{person}{Jeff~Z HaoChen}, \bibinfo{person}{Adrien Gaidon}, {and} \bibinfo{person}{Tengyu Ma}.} \bibinfo{year}{2021}\natexlab{}.
\newblock \showarticletitle{Self-supervised learning is more robust to dataset imbalance}.
\newblock \bibinfo{journal}{\emph{arXiv preprint arXiv:2110.05025}} (\bibinfo{year}{2021}).
\newblock


\bibitem[Liu et~al\mbox{.}(2017a)]%
        {spam}
\bibfield{author}{\bibinfo{person}{Shigang Liu}, \bibinfo{person}{Yu Wang}, \bibinfo{person}{Jun Zhang}, \bibinfo{person}{Chao Chen}, {and} \bibinfo{person}{Yang Xiang}.} \bibinfo{year}{2017}\natexlab{a}.
\newblock \showarticletitle{Addressing the class imbalance problem in twitter spam detection using ensemble learning}.
\newblock \bibinfo{journal}{\emph{Computers \& Security}}  \bibinfo{volume}{69} (\bibinfo{year}{2017}), \bibinfo{pages}{35--49}.
\newblock


\bibitem[Liu et~al\mbox{.}(2017b)]%
        {liu2017sphereface}
\bibfield{author}{\bibinfo{person}{Weiyang Liu}, \bibinfo{person}{Yandong Wen}, \bibinfo{person}{Zhiding Yu}, \bibinfo{person}{Ming Li}, \bibinfo{person}{Bhiksha Raj}, {and} \bibinfo{person}{Le Song}.} \bibinfo{year}{2017}\natexlab{b}.
\newblock \showarticletitle{Sphereface: Deep hypersphere embedding for face recognition}. In \bibinfo{booktitle}{\emph{Proceedings of the IEEE conference on computer vision and pattern recognition}}. \bibinfo{pages}{212--220}.
\newblock


\bibitem[Liu et~al\mbox{.}(2008)]%
        {128}
\bibfield{author}{\bibinfo{person}{Xu-Ying Liu}, \bibinfo{person}{Jianxin Wu}, {and} \bibinfo{person}{Zhi-Hua Zhou}.} \bibinfo{year}{2008}\natexlab{}.
\newblock \showarticletitle{Exploratory undersampling for class-imbalance learning}.
\newblock \bibinfo{journal}{\emph{IEEE Transactions on Systems, Man, and Cybernetics, Part B (Cybernetics)}} \bibinfo{volume}{39}, \bibinfo{number}{2} (\bibinfo{year}{2008}), \bibinfo{pages}{539--550}.
\newblock


\bibitem[Liu et~al\mbox{.}(2019)]%
        {places}
\bibfield{author}{\bibinfo{person}{Ziwei Liu}, \bibinfo{person}{Zhongqi Miao}, \bibinfo{person}{Xiaohang Zhan}, \bibinfo{person}{Jiayun Wang}, \bibinfo{person}{Boqing Gong}, {and} \bibinfo{person}{Stella~X Yu}.} \bibinfo{year}{2019}\natexlab{}.
\newblock \showarticletitle{Large-scale long-tailed recognition in an open world}. In \bibinfo{booktitle}{\emph{Proceedings of the IEEE/CVF conference on computer vision and pattern recognition}}. \bibinfo{pages}{2537--2546}.
\newblock


\bibitem[Mahajan et~al\mbox{.}(2018)]%
        {root}
\bibfield{author}{\bibinfo{person}{Dhruv Mahajan} {et~al\mbox{.}}} \bibinfo{year}{2018}\natexlab{}.
\newblock \showarticletitle{Exploring the limits of weakly supervised pretraining}. In \bibinfo{booktitle}{\emph{Proceedings of the European conference on computer vision}}.
\newblock


\bibitem[Masko and Hensman(2015)]%
        {survey12}
\bibfield{author}{\bibinfo{person}{David Masko} {and} \bibinfo{person}{Paulina Hensman}.} \bibinfo{year}{2015}\natexlab{}.
\newblock \bibinfo{title}{The impact of imbalanced training data for convolutional neural networks}.
\newblock
\newblock


\bibitem[Menon et~al\mbox{.}(2013)]%
        {menon2013statistical}
\bibfield{author}{\bibinfo{person}{Aditya Menon}, \bibinfo{person}{Harikrishna Narasimhan}, \bibinfo{person}{Shivani Agarwal}, {and} \bibinfo{person}{Sanjay Chawla}.} \bibinfo{year}{2013}\natexlab{}.
\newblock \showarticletitle{On the statistical consistency of algorithms for binary classification under class imbalance}. In \bibinfo{booktitle}{\emph{International Conference on Machine Learning}}. PMLR, \bibinfo{pages}{603--611}.
\newblock


\bibitem[Menon et~al\mbox{.}(2020)]%
        {menon2020long}
\bibfield{author}{\bibinfo{person}{Aditya~Krishna Menon}, \bibinfo{person}{Sadeep Jayasumana}, \bibinfo{person}{Ankit~Singh Rawat}, \bibinfo{person}{Himanshu Jain}, \bibinfo{person}{Andreas Veit}, {and} \bibinfo{person}{Sanjiv Kumar}.} \bibinfo{year}{2020}\natexlab{}.
\newblock \showarticletitle{Long-tail learning via logit adjustment}.
\newblock \bibinfo{journal}{\emph{arXiv preprint arXiv:2007.07314}} (\bibinfo{year}{2020}).
\newblock


\bibitem[Naeem et~al\mbox{.}({[n.\,d.]})]%
        {tail4}
\bibfield{author}{\bibinfo{person}{Muhammad~Ferjad Naeem}, \bibinfo{person}{Muhammad Gul Zain~Ali Khan}, {et~al\mbox{.}}} \bibinfo{year}{[n.\,d.]}\natexlab{}.
\newblock \showarticletitle{Supplementary-I2MVFormer: Large Language Model Generated Multi-View Document Supervision for Zero-Shot Image Classification}.
\newblock  (\bibinfo{year}{[n.\,d.]}).
\newblock


\bibitem[Peng et~al\mbox{.}(2022)]%
        {peng2022optimal}
\bibfield{author}{\bibinfo{person}{Hanyu Peng}, \bibinfo{person}{Mingming Sun}, {and} \bibinfo{person}{Ping Li}.} \bibinfo{year}{2022}\natexlab{}.
\newblock \showarticletitle{Optimal transport for long-tailed recognition with learnable cost matrix}. In \bibinfo{booktitle}{\emph{International Conference on Learning Representations}}.
\newblock


\bibitem[Perez and Wang(2017)]%
        {150}
\bibfield{author}{\bibinfo{person}{Luis Perez} {and} \bibinfo{person}{Jason Wang}.} \bibinfo{year}{2017}\natexlab{}.
\newblock \showarticletitle{The effectiveness of data augmentation in image classification using deep learning}.
\newblock \bibinfo{journal}{\emph{preprint arXiv:1712.04621}} (\bibinfo{year}{2017}).
\newblock


\bibitem[Platt et~al\mbox{.}(1999)]%
        {platt1999probabilistic}
\bibfield{author}{\bibinfo{person}{John Platt} {et~al\mbox{.}}} \bibinfo{year}{1999}\natexlab{}.
\newblock \showarticletitle{Probabilistic outputs for support vector machines and comparisons to regularized likelihood methods}.
\newblock \bibinfo{journal}{\emph{Advances in large margin classifiers}} \bibinfo{volume}{10}, \bibinfo{number}{3} (\bibinfo{year}{1999}), \bibinfo{pages}{61--74}.
\newblock


\bibitem[Radford et~al\mbox{.}(2021)]%
        {clip}
\bibfield{author}{\bibinfo{person}{Alec Radford}, \bibinfo{person}{Jong~Wook Kim}, \bibinfo{person}{Chris Hallacy}, \bibinfo{person}{Aditya Ramesh}, \bibinfo{person}{Gabriel Goh}, \bibinfo{person}{Sandhini Agarwal}, \bibinfo{person}{Girish Sastry}, \bibinfo{person}{Amanda Askell}, \bibinfo{person}{Pamela Mishkin}, \bibinfo{person}{Jack Clark}, {et~al\mbox{.}}} \bibinfo{year}{2021}\natexlab{}.
\newblock \showarticletitle{Learning transferable visual models from natural language supervision}. In \bibinfo{booktitle}{\emph{International conference on machine learning}}. PMLR.
\newblock


\bibitem[Rangwani et~al\mbox{.}(2022)]%
        {rangwani2022escaping}
\bibfield{author}{\bibinfo{person}{Harsh Rangwani}, \bibinfo{person}{Sumukh~K Aithal}, \bibinfo{person}{Mayank Mishra}, {and} \bibinfo{person}{R~Venkatesh Babu}.} \bibinfo{year}{2022}\natexlab{}.
\newblock \showarticletitle{Escaping Saddle Points for Effective Generalization on Class-Imbalanced Data}.
\newblock \bibinfo{journal}{\emph{arXiv preprint arXiv:2212.13827}} (\bibinfo{year}{2022}).
\newblock


\bibitem[Ren et~al\mbox{.}(2020)]%
        {ren2020balanced}
\bibfield{author}{\bibinfo{person}{Jiawei Ren}, \bibinfo{person}{Cunjun Yu}, \bibinfo{person}{Xiao Ma}, \bibinfo{person}{Haiyu Zhao}, \bibinfo{person}{Shuai Yi}, {et~al\mbox{.}}} \bibinfo{year}{2020}\natexlab{}.
\newblock \showarticletitle{Balanced meta-softmax for long-tailed visual recognition}.
\newblock \bibinfo{journal}{\emph{Advances in neural information processing systems}}  \bibinfo{volume}{33} (\bibinfo{year}{2020}), \bibinfo{pages}{4175--4186}.
\newblock


\bibitem[Ren et~al\mbox{.}(2018)]%
        {400}
\bibfield{author}{\bibinfo{person}{Mengye Ren}, \bibinfo{person}{Wenyuan Zeng}, \bibinfo{person}{Bin Yang}, {and} \bibinfo{person}{Raquel Urtasun}.} \bibinfo{year}{2018}\natexlab{}.
\newblock \showarticletitle{Learning to reweight examples for robust deep learning}. In \bibinfo{booktitle}{\emph{International conference on machine learning}}. PMLR, \bibinfo{pages}{4334--4343}.
\newblock


\bibitem[Rethmeier and Augenstein(2020)]%
        {tail3}
\bibfield{author}{\bibinfo{person}{Nils Rethmeier} {and} \bibinfo{person}{Isabelle Augenstein}.} \bibinfo{year}{2020}\natexlab{}.
\newblock \showarticletitle{Self-supervised contrastive zero to few-shot learning from small, long-tailed text data}.
\newblock  (\bibinfo{year}{2020}).
\newblock


\bibitem[Rethmeier and Augenstein(2022)]%
        {tail2}
\bibfield{author}{\bibinfo{person}{Nils Rethmeier} {and} \bibinfo{person}{Isabelle Augenstein}.} \bibinfo{year}{2022}\natexlab{}.
\newblock \showarticletitle{Long-tail zero and few-shot learning via contrastive pretraining on and for small data}. In \bibinfo{booktitle}{\emph{Computer Sciences \& Mathematics Forum}}, Vol.~\bibinfo{volume}{3}. MDPI, \bibinfo{pages}{10}.
\newblock


\bibitem[Samuel et~al\mbox{.}(2021)]%
        {tail1}
\bibfield{author}{\bibinfo{person}{Dvir Samuel}, \bibinfo{person}{Yuval Atzmon}, {and} \bibinfo{person}{Gal Chechik}.} \bibinfo{year}{2021}\natexlab{}.
\newblock \showarticletitle{From generalized zero-shot learning to long-tail with class descriptors}. In \bibinfo{booktitle}{\emph{Proceedings of the IEEE/CVF winter conference on applications of computer vision}}. \bibinfo{pages}{286--295}.
\newblock


\bibitem[Shorten and Khoshgoftaar(2019)]%
        {151}
\bibfield{author}{\bibinfo{person}{Connor Shorten} {and} \bibinfo{person}{Taghi~M Khoshgoftaar}.} \bibinfo{year}{2019}\natexlab{}.
\newblock \showarticletitle{A survey on image data augmentation for deep learning}.
\newblock \bibinfo{journal}{\emph{Journal of big data}} \bibinfo{volume}{6}, \bibinfo{number}{1} (\bibinfo{year}{2019}), \bibinfo{pages}{1--48}.
\newblock


\bibitem[Shu et~al\mbox{.}(2019)]%
        {shu2019meta}
\bibfield{author}{\bibinfo{person}{Jun Shu}, \bibinfo{person}{Qi Xie}, \bibinfo{person}{Lixuan Yi}, \bibinfo{person}{Qian Zhao}, \bibinfo{person}{Sanping Zhou}, \bibinfo{person}{Zongben Xu}, {and} \bibinfo{person}{Deyu Meng}.} \bibinfo{year}{2019}\natexlab{}.
\newblock \showarticletitle{Meta-weight-net: Learning an explicit mapping for sample weighting}.
\newblock \bibinfo{journal}{\emph{Advances in neural information processing systems}}  \bibinfo{volume}{32} (\bibinfo{year}{2019}).
\newblock


\bibitem[Tan et~al\mbox{.}(2020)]%
        {tan2020equalization}
\bibfield{author}{\bibinfo{person}{Jingru Tan}, \bibinfo{person}{Changbao Wang}, \bibinfo{person}{Buyu Li}, \bibinfo{person}{Quanquan Li}, \bibinfo{person}{Wanli Ouyang}, \bibinfo{person}{Changqing Yin}, {and} \bibinfo{person}{Junjie Yan}.} \bibinfo{year}{2020}\natexlab{}.
\newblock \showarticletitle{Equalization loss for long-tailed object recognition}. In \bibinfo{booktitle}{\emph{Proceedings of the IEEE/CVF conference on computer vision and pattern recognition}}. \bibinfo{pages}{11662--11671}.
\newblock


\bibitem[Van~Hulse et~al\mbox{.}(2007)]%
        {survey10}
\bibfield{author}{\bibinfo{person}{Jason Van~Hulse}, \bibinfo{person}{Taghi~M Khoshgoftaar}, {and} \bibinfo{person}{Amri Napolitano}.} \bibinfo{year}{2007}\natexlab{}.
\newblock \showarticletitle{Experimental perspectives on learning from imbalanced data}. In \bibinfo{booktitle}{\emph{Proceedings of the 24th international conference on Machine learning}}. \bibinfo{pages}{935--942}.
\newblock


\bibitem[Veropoulos et~al\mbox{.}(1999)]%
        {veropoulos1999controlling}
\bibfield{author}{\bibinfo{person}{Konstantinos Veropoulos}, \bibinfo{person}{Colin Campbell}, \bibinfo{person}{Nello Cristianini}, {et~al\mbox{.}}} \bibinfo{year}{1999}\natexlab{}.
\newblock \showarticletitle{Controlling the sensitivity of support vector machines}. In \bibinfo{booktitle}{\emph{Proceedings of the international joint conference on AI}}, Vol.~\bibinfo{volume}{55}. Stockholm, \bibinfo{pages}{60}.
\newblock


\bibitem[Wang et~al\mbox{.}(2018)]%
        {wang2018additive}
\bibfield{author}{\bibinfo{person}{Feng Wang}, \bibinfo{person}{Jian Cheng}, \bibinfo{person}{Weiyang Liu}, {and} \bibinfo{person}{Haijun Liu}.} \bibinfo{year}{2018}\natexlab{}.
\newblock \showarticletitle{Additive margin softmax for face verification}.
\newblock \bibinfo{journal}{\emph{IEEE Signal Processing Letters}} \bibinfo{volume}{25}, \bibinfo{number}{7} (\bibinfo{year}{2018}).
\newblock


\bibitem[Wang et~al\mbox{.}(2021)]%
        {wang2021seesaw}
\bibfield{author}{\bibinfo{person}{Jiaqi Wang}, \bibinfo{person}{Wenwei Zhang}, \bibinfo{person}{Yuhang Zang}, \bibinfo{person}{Yuhang Cao}, \bibinfo{person}{Jiangmiao Pang}, \bibinfo{person}{Tao Gong}, \bibinfo{person}{Kai Chen}, \bibinfo{person}{Ziwei Liu}, \bibinfo{person}{Chen~Change Loy}, {and} \bibinfo{person}{Dahua Lin}.} \bibinfo{year}{2021}\natexlab{}.
\newblock \showarticletitle{Seesaw loss for long-tailed instance segmentation}. In \bibinfo{booktitle}{\emph{Proceedings of the IEEE/CVF conference on computer vision and pattern recognition}}. \bibinfo{pages}{9695--9704}.
\newblock


\bibitem[Wang et~al\mbox{.}(2020)]%
        {wang2020long}
\bibfield{author}{\bibinfo{person}{Xudong Wang}, \bibinfo{person}{Long Lian}, \bibinfo{person}{Zhongqi Miao}, \bibinfo{person}{Ziwei Liu}, {and} \bibinfo{person}{Stella~X Yu}.} \bibinfo{year}{2020}\natexlab{}.
\newblock \showarticletitle{Long-tailed recognition by routing diverse distribution-aware experts}.
\newblock \bibinfo{journal}{\emph{arXiv preprint arXiv:2010.01809}} (\bibinfo{year}{2020}).
\newblock


\bibitem[Wang et~al\mbox{.}(2019)]%
        {wang2019dynamic}
\bibfield{author}{\bibinfo{person}{Yiru Wang}, \bibinfo{person}{Weihao Gan}, \bibinfo{person}{Jie Yang}, \bibinfo{person}{Wei Wu}, {and} \bibinfo{person}{Junjie Yan}.} \bibinfo{year}{2019}\natexlab{}.
\newblock \showarticletitle{Dynamic curriculum learning for imbalanced data classification}. In \bibinfo{booktitle}{\emph{Proceedings of the IEEE/CVF international conference on computer vision}}. \bibinfo{pages}{5017--5026}.
\newblock


\bibitem[Wang et~al\mbox{.}(2017)]%
        {401}
\bibfield{author}{\bibinfo{person}{Yu-Xiong Wang}, \bibinfo{person}{Deva Ramanan}, {and} \bibinfo{person}{Martial Hebert}.} \bibinfo{year}{2017}\natexlab{}.
\newblock \showarticletitle{Learning to model the tail}.
\newblock \bibinfo{journal}{\emph{Advances in neural information processing systems}}  \bibinfo{volume}{30} (\bibinfo{year}{2017}).
\newblock


\bibitem[Wei et~al\mbox{.}(2018)]%
        {wei}
\bibfield{author}{\bibinfo{person}{Colin Wei}, \bibinfo{person}{Jason Lee}, \bibinfo{person}{Qiang Liu}, {and} \bibinfo{person}{Tengyu Ma}.} \bibinfo{year}{2018}\natexlab{}.
\newblock \showarticletitle{On the margin theory of feedforward neural networks}.
\newblock  (\bibinfo{year}{2018}).
\newblock


\bibitem[Wei et~al\mbox{.}(2021)]%
        {104}
\bibfield{author}{\bibinfo{person}{Tong Wei}, \bibinfo{person}{Jiang-Xin Shi}, \bibinfo{person}{Wei-Wei Tu}, {and} \bibinfo{person}{Yu-Feng Li}.} \bibinfo{year}{2021}\natexlab{}.
\newblock \showarticletitle{Robust long-tailed learning under label noise}.
\newblock \bibinfo{journal}{\emph{arXiv preprint arXiv:2108.11569}} (\bibinfo{year}{2021}).
\newblock


\bibitem[Yang et~al\mbox{.}(2022)]%
        {s1}
\bibfield{author}{\bibinfo{person}{Lu Yang}, \bibinfo{person}{He Jiang}, \bibinfo{person}{Qing Song}, {and} \bibinfo{person}{Jun Guo}.} \bibinfo{year}{2022}\natexlab{}.
\newblock \showarticletitle{A survey on long-tailed visual recognition}.
\newblock \bibinfo{journal}{\emph{International Journal of Computer Vision}} \bibinfo{volume}{130}, \bibinfo{number}{7} (\bibinfo{year}{2022}).
\newblock


\bibitem[Ye et~al\mbox{.}(2020)]%
        {ye2020identifying}
\bibfield{author}{\bibinfo{person}{Han-Jia Ye}, \bibinfo{person}{Hong-You Chen}, \bibinfo{person}{De-Chuan Zhan}, {and} \bibinfo{person}{Wei-Lun Chao}.} \bibinfo{year}{2020}\natexlab{}.
\newblock \showarticletitle{Identifying and compensating for feature deviation in imbalanced deep learning}.
\newblock \bibinfo{journal}{\emph{arXiv preprint arXiv:2001.01385}} (\bibinfo{year}{2020}).
\newblock


\bibitem[Zhang et~al\mbox{.}(2021b)]%
        {zhang2021distribution}
\bibfield{author}{\bibinfo{person}{Songyang Zhang}, \bibinfo{person}{Zeming Li}, \bibinfo{person}{Shipeng Yan}, \bibinfo{person}{Xuming He}, {and} \bibinfo{person}{Jian Sun}.} \bibinfo{year}{2021}\natexlab{b}.
\newblock \showarticletitle{Distribution alignment: A unified framework for long-tail visual recognition}. In \bibinfo{booktitle}{\emph{Proceedings of the IEEE/CVF conference on computer vision and pattern recognition}}. \bibinfo{pages}{2361--2370}.
\newblock


\bibitem[Zhang et~al\mbox{.}(2021a)]%
        {zhang2021self}
\bibfield{author}{\bibinfo{person}{Yifan Zhang}, \bibinfo{person}{Bryan Hooi}, \bibinfo{person}{Lanqing Hong}, {and} \bibinfo{person}{Jiashi Feng}.} \bibinfo{year}{2021}\natexlab{a}.
\newblock \showarticletitle{Self-supervised aggregation of diverse experts for test-agnostic long-tailed recognition}.
\newblock \bibinfo{journal}{\emph{arXiv preprint arXiv:2107.09249}} (\bibinfo{year}{2021}).
\newblock


\bibitem[Zhang et~al\mbox{.}(2023)]%
        {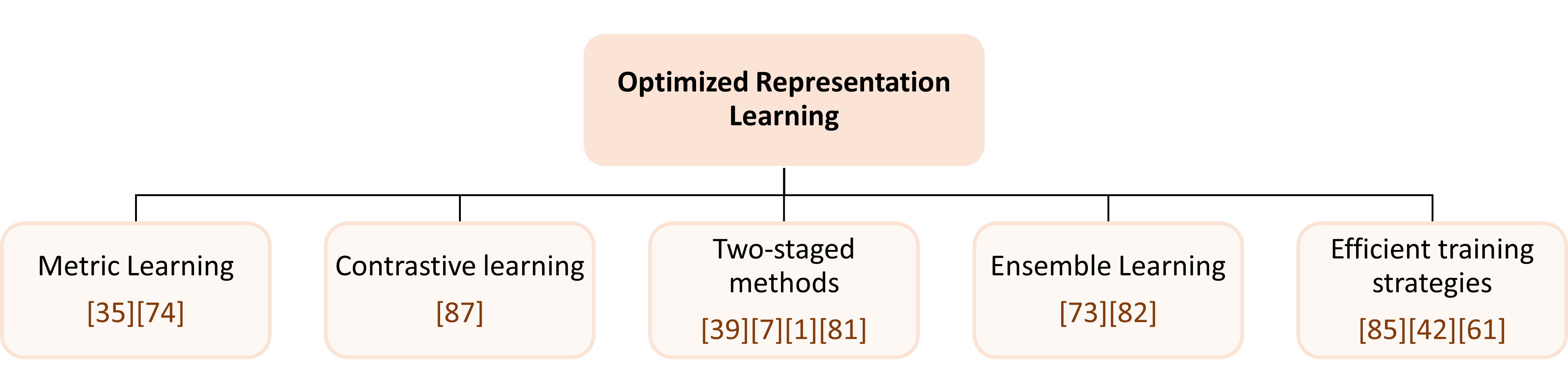}
\bibfield{author}{\bibinfo{person}{Yifan Zhang}, \bibinfo{person}{Bingyi Kang}, \bibinfo{person}{Bryan Hooi}, \bibinfo{person}{Shuicheng Yan}, {and} \bibinfo{person}{Jiashi Feng}.} \bibinfo{year}{2023}\natexlab{}.
\newblock \showarticletitle{Deep long-tailed learning: A survey}.
\newblock \bibinfo{journal}{\emph{IEEE Transactions on Pattern Analysis and Machine Intelligence}} (\bibinfo{year}{2023}).
\newblock


\bibitem[Zhang and Pfister(2021)]%
        {129}
\bibfield{author}{\bibinfo{person}{Zizhao Zhang} {and} \bibinfo{person}{Tomas Pfister}.} \bibinfo{year}{2021}\natexlab{}.
\newblock \showarticletitle{Learning fast sample re-weighting without reward data}. In \bibinfo{booktitle}{\emph{Proceedings of the IEEE/CVF International Conference on Computer Vision}}. \bibinfo{pages}{725--734}.
\newblock


\bibitem[Zhong et~al\mbox{.}(2019)]%
        {zhong2019unequal}
\bibfield{author}{\bibinfo{person}{Yaoyao Zhong}, \bibinfo{person}{Weihong Deng}, \bibinfo{person}{Mei Wang}, \bibinfo{person}{Jiani Hu}, \bibinfo{person}{Jianteng Peng}, \bibinfo{person}{Xunqiang Tao}, {and} \bibinfo{person}{Yaohai Huang}.} \bibinfo{year}{2019}\natexlab{}.
\newblock \showarticletitle{Unequal-training for deep face recognition with long-tailed noisy data}. In \bibinfo{booktitle}{\emph{Proceedings of the IEEE/CVF Conference on Computer Vision and Pattern Recognition}}. \bibinfo{pages}{7812--7821}.
\newblock


\bibitem[Zhou et~al\mbox{.}(2020)]%
        {zhou2020bbn}
\bibfield{author}{\bibinfo{person}{Boyan Zhou}, \bibinfo{person}{Quan Cui}, \bibinfo{person}{Xiu-Shen Wei}, {and} \bibinfo{person}{Zhao-Min Chen}.} \bibinfo{year}{2020}\natexlab{}.
\newblock \showarticletitle{Bbn: Bilateral-branch network with cumulative learning for long-tailed visual recognition}. In \bibinfo{booktitle}{\emph{Proceedings of the IEEE/CVF conference on computer vision and pattern recognition}}. \bibinfo{pages}{9719--9728}.
\newblock


\bibitem[Zhu et~al\mbox{.}(2023)]%
        {zf}
\bibfield{author}{\bibinfo{person}{Beier Zhu}, \bibinfo{person}{Yulei Niu}, \bibinfo{person}{Yucheng Han}, \bibinfo{person}{Yue Wu}, {and} \bibinfo{person}{Hanwang Zhang}.} \bibinfo{year}{2023}\natexlab{}.
\newblock \showarticletitle{Prompt-aligned gradient for prompt tuning}. In \bibinfo{booktitle}{\emph{Proceedings of the IEEE/CVF International Conference on Computer Vision}}. \bibinfo{pages}{15659--15669}.
\newblock


\bibitem[Zhu et~al\mbox{.}(2022)]%
        {zhu2022balanced}
\bibfield{author}{\bibinfo{person}{Jianggang Zhu}, \bibinfo{person}{Zheng Wang}, \bibinfo{person}{Jingjing Chen}, \bibinfo{person}{Yi-Ping~Phoebe Chen}, {and} \bibinfo{person}{Yu-Gang Jiang}.} \bibinfo{year}{2022}\natexlab{}.
\newblock \showarticletitle{Balanced contrastive learning for long-tailed visual recognition}. In \bibinfo{booktitle}{\emph{Proceedings of the IEEE/CVF Conference on Computer Vision and Pattern Recognition}}. \bibinfo{pages}{6908--6917}.
\newblock


\end{thebibliography}
\end{document}